\documentclass[a4paper,fleqn]{cas-dc}
\usepackage{tabularx}
\usepackage{threeparttable}
\usepackage{booktabs} 
\usepackage{graphicx}    
\usepackage{overpic}     
\usepackage{subcaption}  
\usepackage{caption}    
\usepackage[numbers]{natbib}
\usepackage{microtype}
\usepackage{xurl}
\usepackage{hyperref}
\hypersetup{breaklinks=true}
\newtheorem{theorem}{Theorem}

\newtheorem{definition}{Definition}

\makeatletter
\ExplSyntaxOn

\cs_gset_eq:NN \__cas_affiliation_print_orcids: \prg_do_nothing:

\ExplSyntaxOff
\makeatother

\def\tsc#1{\csdef{#1}{\textsc{\lowercase{#1}}\xspace}}
\tsc{WGM}
\tsc{QE}
\tsc{EP}
\tsc{PMS}
\tsc{BEC}
\tsc{DE}

\begin{document}
\let\WriteBookmarks\relax
\def\floatpagepagefraction{1}
\def\textpagefraction{.001}
\shorttitle{TriDP-PTM for Radar Cardiac Sensing}
\shortauthors{J. Li et~al.}

\title [mode = title]{TriDP‑PTM: a three-stage distortion‑perception tradeoff guides the pre-training model for radar cardiac sensing}

\tnotemark[1]

\tnotetext[1]{This work was supported by grants from the National Natural Science Foundation of China (62571009).}


\author[1,2,3]{Jinye Li}[
]
\ead{scae2la@outlook.com}
\credit{Conceptualization, Methodology, Software, Writing - Original Draft Preparation, Visualization}

\author[4]{Aidong Men}[
]
\ead{menad@bupt.edu.cn}
\credit{Validation, Writing - Review \& Editing}

\author[5]{Yang Liu}[
]
\ead{yangliu@pku.edu.cn}
\credit{Resources, Supervision, Writing - Review \& Editing}

\author[1,2,6]{Qingchao Chen}[
]
\cormark[1]
\ead{qingchao.chen@pku.edu.cn}
\credit{Conceptualization, Supervision, Writing - Review \& Editing, Funding Acquisition}

\affiliation[1]{
    organization={National Institute of Health Data Science, Peking University},
    city={Beijing},
    country={China}
}

\affiliation[2]{
    organization={Institute of Medical Technology, Peking University},
    city={Beijing},
    country={China}
}

\affiliation[3]{
    organization={Beijing University of Posts and Telecommunications},
    city={Beijing},
    country={China}
}

\affiliation[4]{
    organization={School of Artificial Intelligence, Beijing University of Posts and Telecommunications},
    city={Beijing},
    country={China}
}

\affiliation[5]{
    organization={Wangxuan Institute of Computer Technology, Peking University},
    city={Beijing},
    country={China}
}

\affiliation[6]{
    organization={State Key Laboratory of General Artificial Intelligence, Peking University},
    city={Beijing},
    country={China}
}

\cortext[cor1]{Corresponding author}





\begin{abstract}
Cardiovascular diseases (CVDs) remain a leading cause of death globally, necessitating continuous, accurate, and non-invasive cardiac monitoring technologies. While non-contact radar-based approaches show great promise, they often employ a single “distortion-driven” or “perception-driven” paradigm, frequently facing a trade-off between “low distortion but weak semantic information” and “high perceptual fidelity but poor interpretability.” To address this, we propose a Three-stage Distortion-Perception Pre-Training Model (TriDP-PTM), which is structured as a radar-based multi-scale fusion dual-path framework that systematically compares the "direct radar-to-task" path against an "indirect radar-to-ECG-to-task" path. By integrating an ECG generator with a feature discriminator to form a composite loss function, our approach effectively incorporates medical priors—such as ECG morphology and rhythm—into downstream tasks. Through empirical analysis, we reveal that this trade-off manifests in three distinct phases (Positive-Sum, Coopetitive, and Negative-Sum), demonstrating that optimal downstream clinical assessment accuracy typically emerges in the coopetitive stage. Extensive experiments on a dataset involving 30 subjects across 5 physiological states reveal that the indirect path consistently outperforms the direct path in diverse tasks, yielding a mean IoU of 0.80 in waveform segmentation, an average classification accuracy of 98.3\% across four classification tasks, and an approximately 56\% MAE reduction in blood pressure regression compared to the strongest baselines. These findings validate our framework and indicate that, within the indirect radar-to-ECG pathway, appropriately weighting the distortion and perception losses to operate in the coopetitive regime is critical for achieving both clinically interpretable ECG morphology and strong downstream accuracy in non-contact cardiac monitoring.

\end{abstract}

\begin{graphicalabstract}
\includegraphics{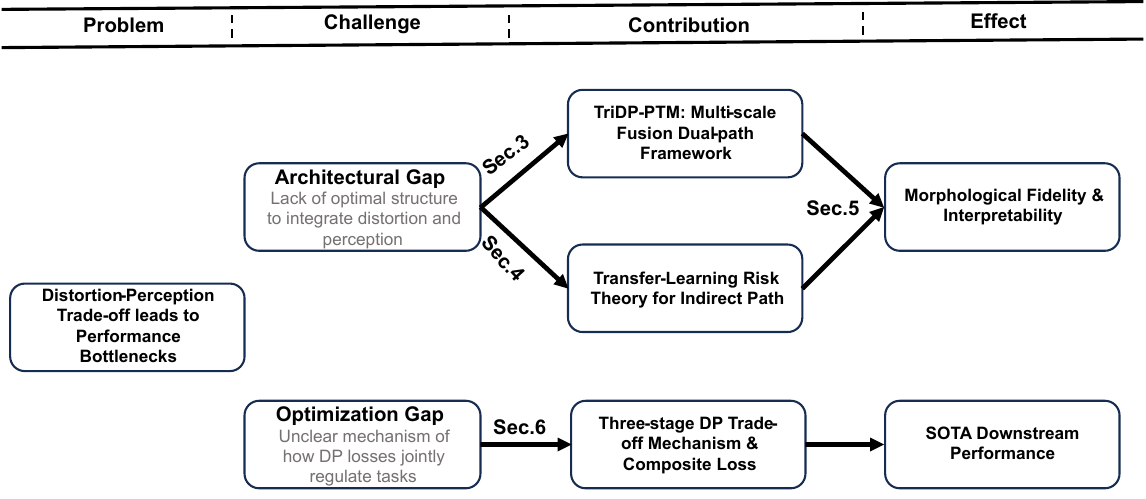}
\end{graphicalabstract}

\begin{highlights}
\item Proposes TriDP-PTM, a multi-scale fusion dual-path pre-training model for contactless cardiac monitoring.
\item Demonstrates that the indirect radar-to-ECG pathway outperforms direct radar-to-task learning across diverse downstream tasks.
\item Identifies a three-stage distortion-perception trade-off, with the coopetitive regime achieving the best fidelity-accuracy balance.
\end{highlights}


\maketitle

\section{Introduction}
\label{sec:introduction}

Cardiovascular diseases (CVDs) persist as a leading cause of mortality worldwide,  highlighting the urgent need for continuous, accurate cardiac monitoring to enable early diagnosis and personalized therapeutic interventions\cite{collaborators2015global}. Although conventional electrocardiogram (ECG) systems are clinically validated for capturing detailed cardiac electrical activity, their reliance on skin-contact electrodes and high patient compliance requirements significantly limit their applicability for long-term monitoring, especially in dynamic environments such as sleep or physical activity\cite{yeo2021respiratory, garcia2024sleepecg, yeo2022robust,  tripathi2022ensemble, al2023adaptive}, or for preventative early screening\cite{zhang2024monitoring, yuan2025atrial}. In recent years, contactless radar sensing technology has emerged as a highly promising alternative for physiological monitoring. By detecting subtle movements of the chest wall \cite{seifizarei2025continuous,xu2024health, kwon2021attention, chen2022contactless, zhang2024monitoring, yuan2025atrial, hu2024investigating, li2024fast, lopes2024covis, marty2024frequency, xu2021accurate, moadi2024enhancing}, this approach enables the estimation of fundamental cardiac physiological parameters without any physical contact, thereby facilitating long-term monitoring of individuals.

\begin{figure*}[h!]
\centering
  \includegraphics[width=1\textwidth]{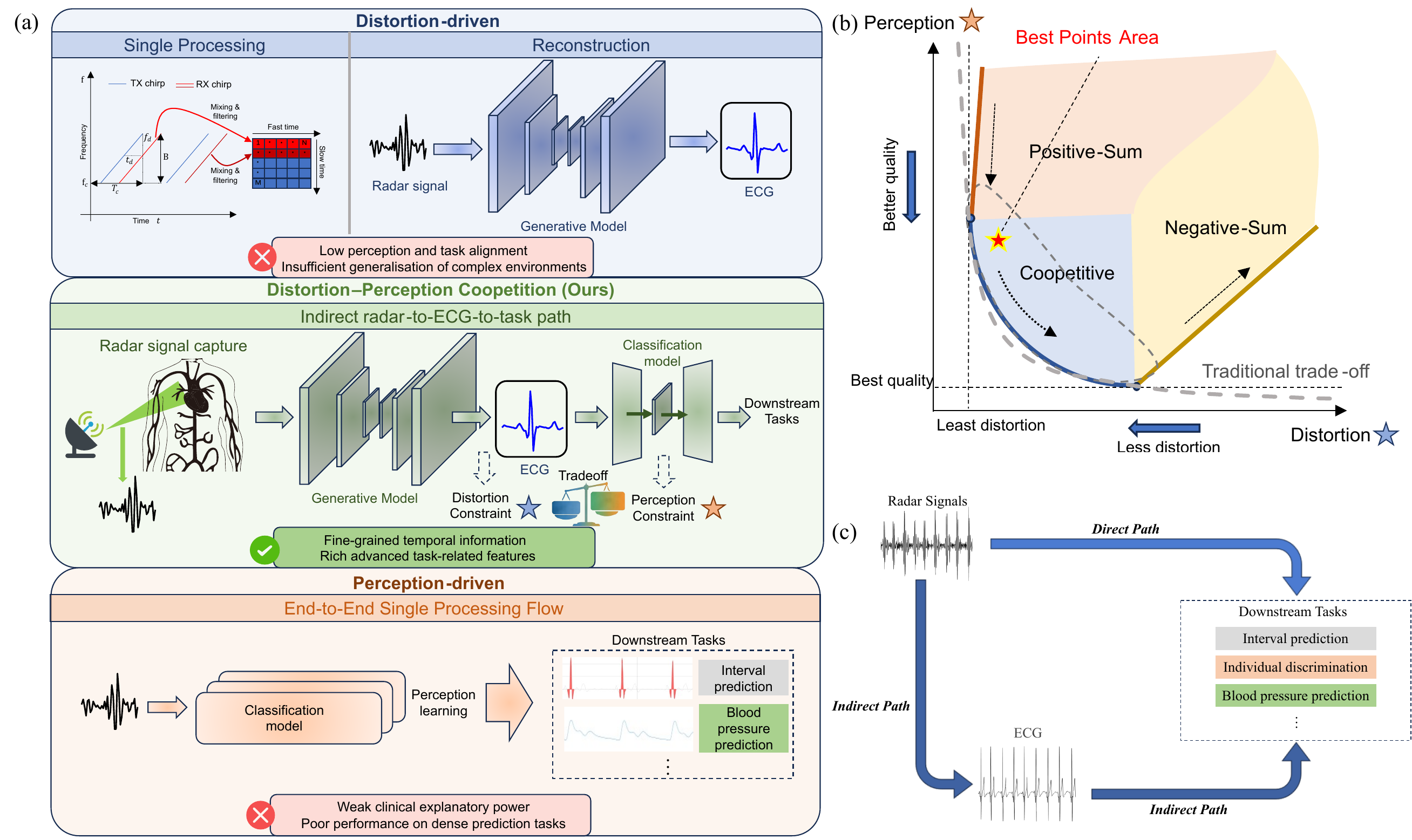}
  \vspace{-3mm} 
  \caption{(a) Contrasts two prevailing paradigms: distortion‑driven and perception‑driven. We optimize downstream performance via a principled trade‑off between distortion (blue stars) and perception (orange stars). (b) Contrasts the conventional monotonic distortion–perception trade-off with the three-stage trade-off observed along the indirect (ECG-mediated) path. Specifically, approaching the minimum-distortion point corresponds to a positive-sum regime; progressing from the minimum-distortion point toward the best-perception point corresponds to a coopetitive regime; and once the trajectory moves away from the best-perception point, it enters a negative-sum regime. Empirically, the optimal operating point for downstream tasks lies within, or near, the coopetitive regime. (c) Illustrates the workflows of the indirect path and direct path: the direct path predicts downstream tasks directly from radar heart sound signals, while the indirect path first converts radar heart sound signals into ECG signals and then predicts downstream tasks from the ECG signals.}
  \vspace{-3mm}
  \label{teaser}
\end{figure*}

Currently, radar-based cardiac sensing methods can be broadly categorized into two modes, as illustrated in Figure \ref{teaser}a. One is \textbf{distortion-driven}, using signal processing to generate ECG-like signals or directly generating more realistic ECGs to assess basic physiological indicators. The other is \textbf{perception-driven}, which directly maps radar signals to physiological indicators and disease predictions. Distortion-driven methods treat the problem as a signal generation task, aiming to "translate" radar signals into waveforms that geometrically match clinical ECGs for subsequent analysis. Typical workflows include: directly estimating heart rate or respiratory rate using filter banks\cite{seifizarei2025continuous,xu2024health, hu2024investigating, li2024fast, lopes2024covis, marty2024frequency, xu2021accurate, moadi2024enhancing,xiang2022high,liu2023mmrh,li2023multidomain,zhang2021mutual,wang2020remote,zheng2022second,qu2023vital,mercuri2019vital}; extracting more complex metrics, such as heart rate variability (HRV), through empirical mode decomposition (EMD) or variational mode decomposition (VMD)\cite{wang2021driver,antolinos2020cardiopulmonary,zhang2024monitoring,wang2021mmhrv}; generating high-fidelity ECGs using deep neural networks\cite{chen2022contactless} to subsequently calculate physiological parameters (e.g., heart rate) from the generated signals. This paradigm incorporates prior clinical information, provides morphological interpretability, and aligns with clinical workflows. However, they exhibit high sensitivity to algorithmic parameters and network fitting, often prioritizing “low distortion” at the expense of “high perceptual quality” and task alignment. Consequently, they typically only achieve stable computation of basic numerical metrics, with insufficient generalisation capabilities in complex radar acquisition environments.

In contrast, perception-driven approaches perform end-to-end inference directly from radar signals to perform downstream tasks, such as estimating heart rate\cite{zhao2020heart, gruzewska2025uwb}, respiration\cite{wang2024rf}, or blood pressure\cite{wang2024uwb}, and detecting arrhythmias
\cite{zhang2024monitoring} and atrial fibrillation\cite{yuan2025atrial}. They are generally more robust to noise and domain shifts and better capture high-level features relevant to the task. However, they are less constrained by the original ECG morphology and fine-grained temporal structure, reducing clinical interpretability and often performing poorly in generating realistic physiological signals and dense prediction tasks.

Optimizing solely for distortion or solely for perception leads to performance bottlenecks—the former sacrificing task perception, and the latter sacrificing clinical morphology. Therefore, a principled trade-off on the distortion-perception (DP) axis is crucial. To effectively achieve this trade-off, we must first determine the optimal architectural strategies that simultaneously encompass distortion and perception, and the trade-off principles governing their interaction. Therefore, we propose two fundamental research questions to guide our study: \textbf{Path superiority}: does the simultaneous incorporation of distortion and perception losses facilitate improved performance on downstream tasks? \textbf{Distortion-perception trade-off}: how do distortion and perception losses jointly regulate downstream task outcomes? To answer these questions, we introduce a dual-path comparison framework and jointly optimize distortion and perception losses along the indirect path. This reveals a three-stage distortion-perception mechanism—positive-sum, coopetitive, and negative-sum. Empirical results show that the optimal operating point for downstream tasks lies within or near the coopetitive stage, thereby improving task metrics while maintaining ECG morphology and clinical interpretability.

To systematically evaluate whether combining distortion loss and perceptual loss can improve downstream task performance, we propose a Three-stage Distortion-Perception Pre-Training Model (TriDP-PTM). As shown in Figure \ref{teaser}c, TriDP-PTM is instantiated as a dual-path comparison framework, where both paths employ the same backbone network architecture, consisting of a heart sound encoder and a multi-scale fusion bottleneck \cite{li2025radar2ecg}. This architectural foundation decomposes the joint time-frequency representation to expand effective feature channels while capturing both low-frequency periodicity (e.g., inter-beat interval) and high-frequency morphological details (e.g., QRS waveform). Based on this structural alignment, the two paths employ different inference strategies: the direct path maps the bottleneck features directly to task predictions through a task-specific head network, while the indirect path feeds these features into a symmetric decoder to first synthesize high-fidelity ECG waveforms, and then performs downstream tasks based on these waveforms. Extensive experiments demonstrate that the indirect path consistently outperforms the direct path across multiple downstream tasks. Meanwhile, we further elucidate the intrinsic mechanism of the dual-path framework using the two-stage empirical risk minimization (ERM) paradigm from transfer learning theory. Consistent with the empirical findings, the direct path is constrained by sparse supervision and limited task coverage, whereas the indirect path trades moderate complexity for “dense supervision and task diversity,” achieving a more favorable balance between the efficiency of transferring shared features and model generalization.

Building upon the observed superiority of the indirect path in the dual-path framework, we further investigate the underlying loss-level mechanisms that govern downstream task performance. To this end, we focus on the indirect pathway and examine how the interplay between distortion loss and perceptual loss shapes both ECG reconstruction fidelity and task generalization. Specifically, we adopt an indirect pathway design that incorporates a feature-level discriminator and introduces a novel composite loss function. The generated ECG and its paired ground-truth ECG are jointly fed into a one-dimensional deep residual network (Net1D)\cite{hong2020holmes} used as a feature extractor; the L2 distance between their feature maps defines a perceptual loss, while the waveform-level L2 loss between the generated and real ECG serves as a distortion loss. These two losses are linearly combined with varying weights to form the new objective. By systematically sweeping the weight ratio, we uncover a tri-stage trade-off—positive-sum, coopetitive, and negative-sum phases as illustrated in Figure \ref{teaser}b. Under the conventional distortion–perception paradigm, the two objectives are largely antagonistic and cannot be jointly optimized. In contrast, we discovered a new three-stage DP mechanism, including that: (i) in the positive-sum phase, distortion and perceptual fidelity improve in concert; (ii) as the model approaches the task optimum, a coopetitive phase emerges, in which gains in one dimension require proportional concessions in the other while maintaining an aligned objective; and (iii) when the conflict intensifies, a negative-sum phase occurs, where over-optimizing one dimension ultimately harms both, driving performance away from the target. Notably, the performance optimum for downstream tasks often resides in the vicinity of the coopetitive phase, where a more favorable joint balance between fidelity and perceptual quality can be achieved. This observation is consistent with theoretical insights from multi-objective optimization and from perceptual signal processing.

In summary, the main contributions of this work are as follows:
\begin{itemize}
\item We propose TriDP-PTM, a novel pre-training model for contactless cardiac monitoring. By employing a multi-scale fusion dual-path framework, it explicitly isolates and contrasts "direct radar-to-task" regression against "indirect radar-to-ECG-to-task" synthesis, establishing a robust structural foundation for physiological signal representation.
\item By invoking transfer-learning risk theory, we mathematically formalize and empirically validate the superiority of the indirect pathway. We demonstrate that dense supervision from ECG reconstruction expands task diversity and significantly reduces sample complexity, yielding superior performance across diverse downstream applications.
\item We uncover a three-stage distortion-perception (DP) trade-off mechanism (Positive-Sum, Coopetitive, and Negative-Sum) within the indirect pathway. By introducing a feature-level discriminator and a composite loss, we identify that the optimal operating point for downstream task accuracy fundamentally resides within the coopetitive regime, ensuring both morphological fidelity and semantic interpretability.
\end{itemize}

\section{Related Works}

\subsection{Radar-based heart monitoring}

Non-contact monitoring of cardiac activities and vital signs is critical for early screening of cardiovascular diseases and long-term health management, with radar technology becoming a core solution due to its ability to sense micro-motions and penetrate materials. A method based on contactless radio monitoring and knowledge transfer was proposed for atrial fibrillation (AF) detection, integrating clinical diagnostic expertise to identify AF-related motion patterns\cite{yuan2025atrial}. A millimeter-wave radar system was developed for contactless electrocardiogram (ECG) monitoring, which extracts cardiac mechanical motions and maps them to ECG signals using deep learning\cite{chen2022contactless}. Contactless radio frequency signals were utilized for long-term cardiac activity monitoring, enabling continuous tracking of physiological parameters through optimized signal processing \cite{zhang2024monitoring}. A fast and efficient FMCW radar phase extraction technique at ultra-narrow range was designed for vital sign detection, enhancing distance resolution to mitigate interference \cite{li2024fast}. A method based on higher-order harmonics peak selection was combined with radar non-contact sensors to achieve accurate heart rate and respiration rate detection \cite{xu2021accurate}. An attention-based LSTM network was paired with IR-UWB radar data to realize non-contact sleep stage classification, leveraging temporal dependencies of physiological signals \cite{kwon2021attention}.

Radar-based non-contact monitoring techniques have been further refined for specific scenarios, improving adaptability and multi-target performance. A continuous radar-based heart rate monitoring method using an autocorrelation-based algorithm was applied in intensive care units (ICUs), reducing motion artifacts through channel selection and post-processing \cite{seifizarei2025continuous}. The CoViS contactless health monitoring system was developed for nursing homes, fusing Doppler radar and IR thermal imaging to monitor heart rate, respiration rate, and body temperature \cite{lopes2024covis}. The generalized sidelobe canceller was utilized to enhance multi-subject vital sign estimation, optimizing accuracy in scenarios with multiple targets \cite{moadi2024enhancing}. A comparative analysis of low-power FMCW radars was conducted to explore how frequency selection affects vital sign monitoring performance \cite{marty2024frequency}. The Health-Radar system achieved non-contact multitarget heart rate variability detection using FMCW radar, supporting simultaneous monitoring of multiple subjects \cite{xu2024health}. The mmHRV system realized contactless heart rate variability monitoring using millimeter-wave radio, adapting to long-term health management needs \cite{wang2021mmhrv}. The mmRH method enabled noncontact vital sign detection with an FMCW mm-wave radar, expanding the application scope of radar in vital sign monitoring \cite{liu2023mmrh}.

\subsection{Distortion-perception trade-off}

Balancing perceptual quality and distortion is crucial for optimizing performance in spatiotemporal forecasting and signal processing. Recent studies have advanced both theoretical frameworks and practical solutions for this tradeoff using diverse analytical and computational approaches. Yoosuf et al. addressed the blurriness issue in MSE-trained spatiotemporal forecasting models by proposing total variation-based regularization, effectively enhancing perceptual quality while maintaining competitive distortion metrics \cite{yoosuf2024improving}. Freirich et al. developed a foundational theory for the distortion-perception tradeoff in Wasserstein space, deriving closed-form expressions for the tradeoff function and optimal estimator interpolation formulas\cite{freirich2021theory}.

Qu et al. established a rate-distortion-perception framework for Gaussian vector sources, providing single-letter characterizations and reverse water-filling solutions for quadratic Wasserstein metrics\cite{qu2025rate}. Furthermore, Salehkalaibar et al. investigated conditional-distribution-based perception measures, deriving discrete source tradeoff characterizations and revealing a 3dB distortion penalty for perfect perceptual quality\cite{salehkalaibar2024rate}. Blau and Michaeli formalized the inherent conflict between distortion and perceptual quality, laying theoretical groundwork for subsequent tradeoff analyses\cite{blau2018perception}. Additionally, Hamdi et al. explored the role of private randomness in rate-distortion-perception tradeoffs, showing its ineffectiveness for compression rates below source entropy\cite{hamdi2024rate}.

\section{Method}

\begin{figure*}[h!]
\centering
  \includegraphics[width=1\textwidth]{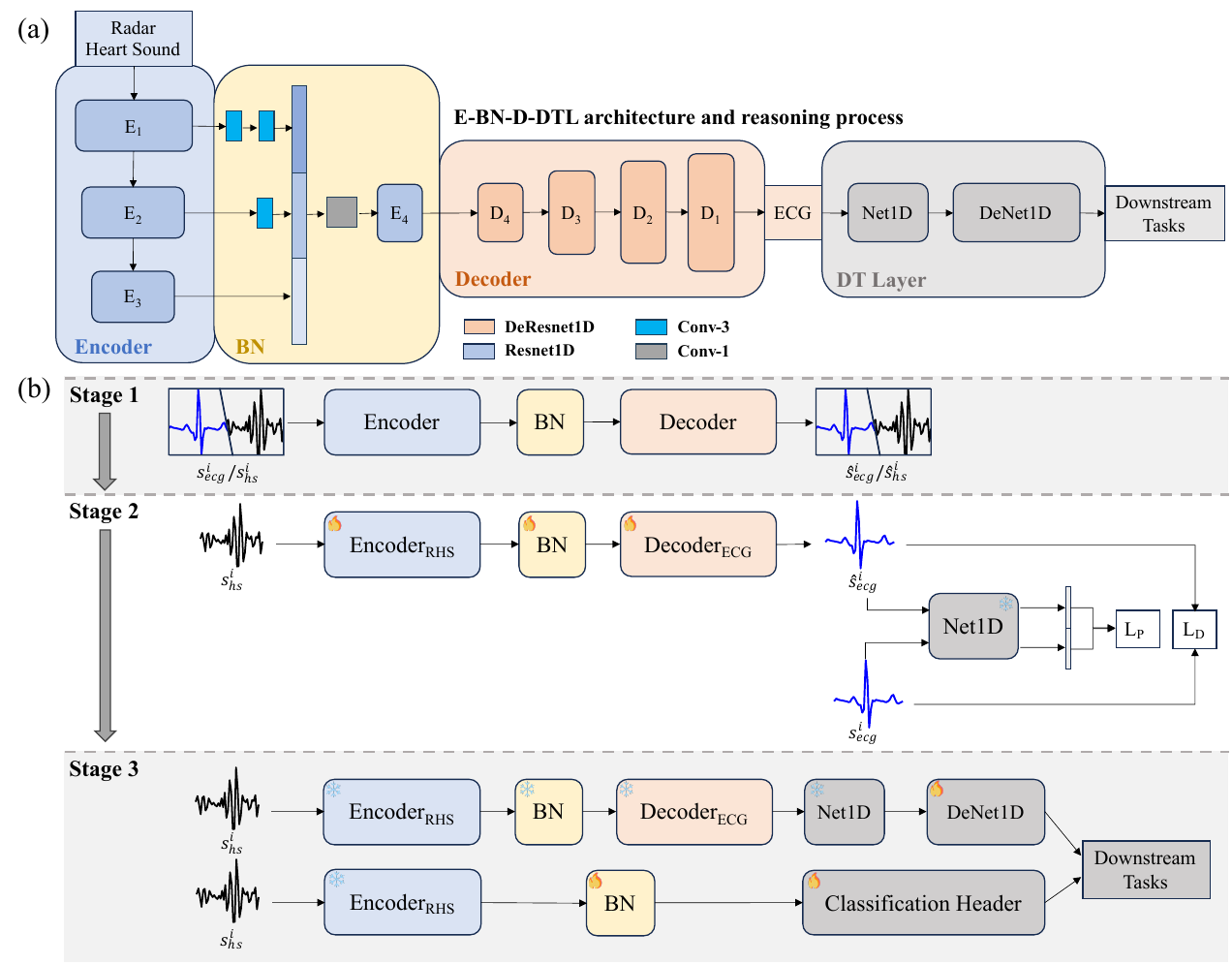}
  \vspace{-3mm} 
  \caption{Framework Overview: (a) illustrates the encoder-bottleneck-decoder-downstream task layer architecture, where the bottleneck aligns multi-scale features from different encoder layers to compress features, eliminate redundancy, and enhance decoding. The downstream task layer features a symmetric autoencoder structure that can be adjusted for various downstream tasks. (b) outlines the three training phases: pre-training (where the model generates its own signals), base task training (generating ECG from radar heart sound signals), and downstream task-specific training (predicting downstream tasks using radar heart sound signals and ECG signals).}
  \vspace{-3mm}
  \label{method}
\end{figure*}

\subsection{Overall}

Current research predominantly targets either translating radar signals into ECG or performing downstream tasks (e.g., segmentation and classification) directly from radar inputs, while underappreciating that the medical priors learned in the generative pathway can more effectively support clinical downstream objectives. To systematically compare the direct radar-to-task mapping against an indirect, ECG-mediated route—and to anchor training in physiologically interpretable supervision—we adopt a dual-path comparative framework equipped with a multi-scale fusion bottleneck and a three-stage training protocol. This design leverages dense ECG supervision while maintaining compact, transferable representations, thereby enabling robust downstream estimation.

\noindent \textbf{Overview of the method design} The proposed Three-stage Distortion-Perception Pre-Training Model (TriDP-PTM) operates through a dual‑path comparison framework, as illustrated in Figure \ref{method}. Both the direct and indirect paths share an encoder and a multi‑scale fusion bottleneck module. In our design, the conventional single output of the encoder’s final layer is replaced by this multi‑scale fusion bottleneck, wherein feature maps from multiple encoder layers are first aligned and then fused to compress the representation, effectively eliminating redundancy and thereby better capturing the complex characteristics of heart sound feature extraction. The two paths diverge in their decoding strategies: the direct path feeds its fused features straight into the downstream task decoder, whereas the indirect path routes them through a decoder whose architecture mirrors that of the encoder to reconstruct the corresponding ECG signal, with an auxiliary feature discriminator imposed to provide perceptual supervision. The overall training protocol is carried out in three sequential phases: \textit{(a) the pre‑training stage, (b) the base‑task training stage, and (c) the downstream‑task fine‑tuning stage.}

\noindent \textbf{Problem Formulation} Let $P=\{(s^{i}_{hs},s^{i}_{ecg})\}_{i=1}^{n}$ represent the training dataset consisting of paired heart sound signals and ECG signals, where $s^{i}_{hs}$ represents the heart sound signal in the $i$-th signal pair, $s^{i}_{ecg}$ represents the corresponding ECG signal in the same pair and $n$ denotes the total number of signal pairs. \textbf{In the pre-training stage}, autoencoders for both heart sound and ECG signals are trained to ensure that the self-generated signal $\hat{S}_{ecg}$ closely resembles the original signal $s_{ecg} \in P$, effectively preparing the encoder and decoder to capture and store a rich representation of features for the subsequent training stage. \textbf{In the base‑task training stage}, only the indirect path is optimized. Leveraging the pre‑trained heart sound encoder and the pre‑trained ECG decoder, we balance the distortion loss—defined as the distance between the original ECG signal and the reconstructed ECG signal—and the perceptual loss—defined as the distance between their feature representations—by adjusting their relative weighting so as to reach an optimal trade‑off in the overall objective. \textbf{In the downstream‑task fine‑tuning stage}, the direct path employs the pre‑trained heart sound encoder while training its downstream‑task decoder; concurrently, the indirect path uses the feature discriminator as its encoder to train its corresponding downstream‑task decoder.

\subsection{DP-Inspired Feature Discriminator}

Within the encoder–multi‑scale fusion bottleneck–decoder architecture, using the reconstruction loss—i.e., the distance between the decoder’s output and the ground truth ECG—as the sole optimization objective provides only a limited perspective on signal quality. This distortion-based loss quantifies numerical discrepancies between the generated and reference waveforms, but fails to capture the fundamental physiological characteristics that define an ECG signal. These include the morphological integrity of the QRS complex, the baseline stability of the ST segment, and the rhythmic periodicity of cardiac cycles—all of which reflect the intrinsic statistical properties of human cardiac activity. Overemphasis on minimizing distortion may yield ECG signals that are numerically close to the ground truth yet suffer from degraded clinical quality, manifesting as blurred waveforms, artificial artifacts, or irregular rhythms. Such deficiencies compromise diagnostic interpretability, as they obscure precisely the features that are critical for medical assessment. The absence of these characteristics renders the generated signals clinically unreliable, regardless of their apparent numerical accuracy.

This limitation can be theoretically grounded in the perception–distortion tradeoff framework, which has been substantiated through mathematical derivation and empirical validation. The theory posits a fundamental inverse relationship between distortion minimization and perceptual quality—defined as the degree to which generated signals adhere to the natural statistics of the data domain. In the context of ECG synthesis, this tradeoff implies that optimizing exclusively for low distortion inevitably shifts the generated signal distribution away from the physiological manifold, resulting in outputs that may be numerically "accurate" but biologically implausible. This phenomenon is not unique to ECG generation but represents a general principle of signal processing manifesting in biomedical applications. Therefore, to produce clinically meaningful ECG signals, it is essential to incorporate perceptual guidance that ensures physiological plausibility within acceptable levels of distortion.

To address this challenge, we enhance the network with a DP-inspired Feature Discriminator, which explicitly encourages the preservation of discriminative ECG features beyond raw waveform similarity. This design ensures that the model remains sensitive to both low-level fidelity and high-level temporal structures essential for clinical and downstream analytical tasks. The Feature Discriminator is implemented using a pre-trained Net1D architecture, which has demonstrated strong effectiveness in extracting hierarchical, discriminative representations from physiological time series. After the decoder reconstructs the ECG waveform, both the generated signal and the reference ECG are passed through the Feature Discriminator to obtain multi-layer feature embeddings. In this setup, the discriminator functions as a perceptual feature extractor that supervises the model to recover rich structural patterns—such as rhythm contours and detailed waveform morphology—that are often overlooked by traditional loss functions.

Working in synergy with the encoder–decoder pipeline, the Feature Discriminator reinforces feature-level consistency, while the multi-scale fusion bottleneck ensures that the decoder receives spatially aligned, compressed representations. This complementary integration fosters robust separation of relevant cardiac dynamics from noise and enhances both the interpretability and fidelity of radar-derived ECG synthesis.

\subsection{Multi-scale Fusion Architecture}

Superior downstream performance hinges on higher-quality ECG reconstruction; radar heart sound-to-ECG synthesis must jointly capture low-frequency periodic structure and high-frequency morphological details. A multi-scale fusion bottleneck aligns and compresses features across encoder depths, reduces redundancy, and mitigates cross-layer misalignment; compared with simple skip connections, it improves reconstruction fidelity and the transferability of learned representations.

Accordingly, in the design of the indirect‐path network architecture, we adopt an encoder–multi‑scale fusion bottleneck–decoder paradigm. The encoder comprises the first three residual blocks of a 1D ResNet, enabling the network to learn progressively deeper hierarchical abstractions of the input heart sound. The decoder, by contrast, consists of all stages of a 1D DeResNet whose structure mirrors that of the encoder, thereby facilitating symmetric feature transformations. Unlike conventional autoencoder and GAN‑based encoder–decoder schemes—which feed only the final encoder activation into the decoder and rely on deep layers to recover global representations—our framework explicitly preserves information from early encoder stages while retaining high‑level context and fine‑grained detail.

To bridge these modules, we introduce a multi‑scale fusion bottleneck that aligns, compresses, and fuses feature maps from different encoder depths before they enter the decoder. Shallow feature maps are progressively downsampled via one‑dimensional convolutional layers to match the spatial resolution of deeper representations; these downsampled maps are then concatenated and condensed by a point‑wise convolution to eliminate redundancy and emphasize correspondences between the heart sound and ECG modalities. The fused output is subsequently injected into the fourth residual block of the ResNet backbone, which serves as the decoder’s input. By performing explicit alignment and fusion at this juncture—instead of relying on U‑Net–style skip connections at every level—our bottleneck effectively corrects misalignments and prevents the propagation of incoherent features into the decoding pathway, thereby enhancing reconstruction fidelity.

\subsection{Optimizations}

In the training process, we perform optimization sequentially in three stages.

Let $P = \{(s^{i}_{hs}, s^{i}_{ecg})\}_{i=1}^{n}$ denote the training dataset consisting of paired heart sound and ECG signals, where $s^{i}_{hs}$ represents the heart sound signal in the $i$-th pair, $s^{i}_{ecg}$ represents the corresponding ECG signal in the same pair, and $n$ is the total number of signal pairs. \textbf{In the pre-training phase}, autoencoders for heart sound and ECG signals are trained to ensure that the self-generated signal $\hat{S}_{ecg}$ is very similar to the original signal $s_{ecg} \in P$, thereby effectively preparing the encoder and decoder to capture and store rich feature representations for use in subsequent training phases. The formula is expressed as follows:

For the heart sound autoencoder, the reconstruction loss is defined as:

\vspace{-1.0ex}
{\setlength\abovedisplayskip{1pt}
\setlength\belowdisplayskip{1pt}
\begin{equation}
\begin{aligned}
\mathcal{L}_{hs} = \frac{1}{n} \sum_{i=1}^{n} \| s^{i}_{hs} - \hat{s}^{i}_{hs} \|_2^2 
\end{aligned}
\label{eq:recon_hs}
\end{equation}}
\vspace{-1.0ex}

where $\hat{s}^{i}_{hs}$ is the reconstructed heart sound signal.

For the ECG autoencoder, the reconstruction loss is defined as:
{\setlength\abovedisplayskip{1pt}
\setlength\belowdisplayskip{1pt}
\begin{equation}
\begin{aligned}
\mathcal{L}_{ecg} = \frac{1}{n} \sum_{i=1}^{n} \| s^{i}_{ecg} - \hat{s}^{i}_{ecg} \|_2^2
\end{aligned}
\label{eq:recon_ecg}
\end{equation}}
\vspace{-1.0ex}

where $\hat{s}^{i}_{ecg}$ is the reconstructed ECG signal.

\textbf{In the base task training phase}, we specifically focus on the theory of perception-distortion tradeoff. To this end, we design two complementary loss functions that compose the main loss function.

The first loss function is the Distortion Loss ($L_d$). Given a signal pair $(s_{hs}, s_{ecg}) \in \{(s^{i}_{hs}, s^{i}_{ecg})\}_{i=1}^{n}$, the projection of the radar heart sound signal in the bottleneck structure is denoted as $\phi$, then the generated ECG is related to the corresponding radar heart sound signal as $\hat{s}_{ecg} = D_{ecg}(\phi)$, where $\phi = BN(E_{hs}(s_{hs}))$, $E_{hs}(\cdot)$ is the radar heart sound pre-training encoder, $BN(\cdot)$ is the bottleneck structure and $D_{ecg}(\cdot)$ is the ECG pre-training decoder. By using the L2 norm to quantify the pixel-level differences between the ECG generated from the heart sound and the original signal, ensuring numerical fidelity, it is defined as:

\vspace{-1.0ex}
{\setlength\abovedisplayskip{1pt}
\setlength\belowdisplayskip{1pt}
\begin{equation}
\begin{aligned}
L_d = \frac{1}{n} \sum_{i=1}^{n} \| s^{i}_{ecg} - \hat{s}^{i}_{ecg} \|_2^2
\end{aligned}
\label{eq:loss_d}
\end{equation}}
\vspace{-1.0ex}

The second loss function is the Perceptual Loss ($L_p$). For the generated signal pair $\{s_{ecg}, \hat{s}_{ecg}\}$, the paired relationship in the feature discriminator is $\{f_{ecg} = M(s_{ecg}), \hat{f}_{ecg} = M(\hat{s}_{ecg})\}$, where $f_{ecg}, \hat{f}_{ecg} \in \mathbb{R}^{C \times L}$, $M(\cdot)$ is the pre-trained feature discriminator (Net1D), $f_{ecg}$ and $\hat{f}_{ecg}$ represent the feature maps output by the last layer of the model for the original ECG signal and the generated ECG signal, respectively, and $C$ and $L$ denote the number of channels and sequence length of the activation features, respectively. By computing the L2 distance of the last-layer feature embeddings extracted by the pre-trained feature discriminator (Net1D), it emphasizes the high-order statistical characteristics of the signal, such as the natural distribution of waveform morphology and the physiological reasonableness of rhythms. It is defined as:

\vspace{-1.0ex}
{\setlength\abovedisplayskip{1pt}
\setlength\belowdisplayskip{1pt}
\begin{equation}
\begin{aligned}
L_p = \frac{1}{n} \sum_{i=1}^{n} \| f_{ecg}^{i} - \hat{f}_{ecg}^{i} \|_2^2
\end{aligned}
\label{eq:loss_p}
\end{equation}}
\vspace{-1.0ex}

This dual-loss design allows us to find the optimal balance point on the DP tradeoff curve: the total loss function is defined as a weighted combination, with individual control parameters before each sub-loss function:

\vspace{-1.0ex}
{\setlength\abovedisplayskip{1pt}
\setlength\belowdisplayskip{1pt}
\begin{equation}
\begin{aligned}
L = \lambda_d L_d + \lambda_p L_p 
\end{aligned}
\label{eq:loss_total}
\end{equation}}
\vspace{-1.0ex}

where $\lambda_d$ and $\lambda_p$ are hyperparameters that control the weights of the distortion loss and perceptual loss, respectively.

\textbf{In the downstream task-specific training phase}, we utilize radar heart sound signals and ECG signals for the direct path and indirect path, respectively, to accomplish downstream tasks.

In the direct path, we employ the pre-trained radar heart sound encoder and bottleneck structure, along with an adjustable downstream task classification head, to predict the downstream tasks. The formula is defined as follows:

\vspace{-1.0ex}
{\setlength\abovedisplayskip{1pt}
\setlength\belowdisplayskip{1pt}
\begin{equation}
\begin{aligned}
\hat{y} = C(\phi), \quad \phi = BN(E_{hs}(s_{hs}))
\end{aligned}
\label{eq:direct_path}
\end{equation}}
\vspace{-1.0ex}

where $E_{hs}(\cdot)$ is the pre-trained encoder, $BN(\cdot)$ is the bottleneck structure, $C(\cdot)$ is the adjustable classification head, $s_{hs}$ is the input radar heart sound signal, and $\hat{y}$ is the predicted output for the downstream task.

In the indirect path, we use Net1D along with DeNet1D, which has an inverse structure to Net1D and is adjustable, to predict the downstream tasks. The formula is defined as follows:

\vspace{-1.0ex}
{\setlength\abovedisplayskip{1pt}
\setlength\belowdisplayskip{1pt}
\begin{equation}
\begin{aligned}
\hat{y} = DeNet1D(Net1D(s_{ecg})) 
\end{aligned}
\label{eq:indirect_path}
\end{equation}}
\vspace{-1.0ex}

where $Net1D(\cdot)$ is the feature extractor, $DeNet1D(\cdot)$ is the adjustable inverse structure.

\section{Theoretical Analysis of Path Comparison}

\subsection{Notation}
We use bold lower-case letters (e.g., $\mathbf{x}$) to denote vectors and bold upper-case letters (e.g., $\mathbf{X}$) to denote matrices. The norm $\|\cdot\|$ applied to a vector or matrix refers to the $\ell_2$ norm or spectral (operator) norm, respectively. We use the bracketed notation $[n]:=\{1,\ldots,n\}$ as shorthand for integer index sets. We use ``hatted'' symbols (e.g., $\hat{\alpha}$, $\hat{B}$) to denote (random) estimators of their underlying population quantities. For a matrix $A$ of rank $r$, $\sigma_1(A),\ldots,\sigma_r(A)$ denote its singular values in decreasing order. Throughout, $\mathcal{F}$ denotes a class of task-specific functions mapping $\mathbb{R}^r\to\mathbb{R}$, and $\mathcal{H}$ denotes a class of shared-representation functions mapping $\mathbb{R}^d\to\mathbb{R}^r$. For the function class $\mathcal{F}$, we write $\mathcal{F}^{\otimes t}$ for its $t$-fold Cartesian product, i.e., $\mathcal{F}^{\otimes t}:=\left\{f\equiv(f_1,\ldots,f_t)\;\middle|\; f_j\in\mathcal{F}\text{ for any }j\in[t]\right\}.$ We use $\tilde{O}(\cdot)$ to denote an expression that hides polylogarithmic factors in problem parameters.

Let $j\in[t]$ index training tasks, each with $n$ samples $\{(x_{ji},y_{ji})\}_{i=1}^n$, and let the target task be indexed by $0$ with $m$ samples $\{(x_{0i},y_{0i})\}_{i=1}^m$. Shared representations are $h\in\mathcal{H}$, task-specific predictors are $f_j\in\mathcal{F}$, and the loss is $\ell:\mathcal{Y}\times\mathcal{Y}\to\mathbb{R}_{\ge 0}$. We use $C(\mathcal{H})$ and $C(\mathcal{F})$ to denote the statistical capacities of the shared-representation and task-specific function classes, and quantify task diversity by $\nu$, reflecting the coverage of the optimal shared representation’s feature space $h^{\ast}$. The direct path is parameterized by $(n_{\mathrm{dir}},t_{\mathrm{dir}},\nu_{\mathrm{dir}},C(\mathcal{F}_{\mathrm{dir}}))$, and the indirect path by $(n_{\mathrm{ind}},t_{\mathrm{ind}},\nu_{\mathrm{ind}},C(\mathcal{F}_{\mathrm{ind}}))$. The training stage may include self-supervised pretraining matched to the data modality (e.g., radar heart sounds or ECG), yielding a learned shared representation $\hat{h}$ via empirical risk minimization over $h\in\mathcal{H}$ and $f\in\mathcal{F}^{\otimes t}$.

\subsection{Transfer Risk Scaling and Indirect Path Dominance}
To anchor the main result, we first recall the empirical training and testing risks in our two-stage ERM setting. For $t$ training tasks (each containing $n$ samples), the cross-task average training risk is

{\setlength\abovedisplayskip{1pt}
\setlength\belowdisplayskip{1pt}
\begin{equation}
\hat{R}_{\mathrm{train}}(f,h):=\frac{1}{nt}\sum_{j=1}^t\sum_{i=1}^n\ell\big(f_j\circ h(x_{ji}),y_{ji}\big),
\label{eq:train-risk}
\end{equation}}

where $f=(f_1,\ldots,f_t)\in\mathcal{F}^{\otimes t}$ collects task-specific predictors and $h\in\mathcal{H}$ is a shared representation. For the target task (indexed by $0$) with $m$ samples, the empirical testing risk is

{\setlength\abovedisplayskip{1pt}
\setlength\belowdisplayskip{1pt}
\begin{equation}
\hat{R}_{\mathrm{test}}(f,h):=\frac{1}{m}\sum_{i=1}^m\ell\big(f\circ h(x_{0i}),y_{0i}\big),
\label{eq:test-risk}
\end{equation}}

with population counterpart $R_{\mathrm{test}}(f,h):=\mathbb{E}[\hat{R}_{\mathrm{test}}(f,h)]$. The Transfer Learning Risk (TLRisk), measuring the deviation from the oracle model on the target task, is defined by

{\setlength\abovedisplayskip{1pt}
\setlength\belowdisplayskip{1pt}
\begin{equation}
\mathrm{TLRisk} := R_{\mathrm{test}}(\hat{f}_0,\hat{h})-R_{\mathrm{test}}(f_0^\ast,h^\ast).
\label{eq:tlrisk-def}
\end{equation}}

Under standard transfer learning assumptions, TLRisk admits the following scaling law:

{\setlength\abovedisplayskip{1pt}
\setlength\belowdisplayskip{1pt}
\begin{equation}
\mathrm{TLRisk}=\tilde{O}\!\left(\frac{1}{\nu}\sqrt{\frac{C(\mathcal{H})+t\,C(\mathcal{F})}{n\,t}}+\sqrt{\frac{C(\mathcal{F})}{m}}\right).
\label{eq:TLRiskScale}
\end{equation}}

This expression shows that transfer learning risk decreases as the training sample sizes $n$ and $t$ grow and as task diversity $\nu$ increases. Crucially, the sample complexity on the target task is governed solely by $C(\mathcal{F})$, and is substantially lower than isolated learning that depends on the composite class $C(\mathcal{F}\circ\mathcal{H})$.

By the scaling law in \eqref{eq:TLRiskScale}, assuming the target task predictor structure is fixed, the adaptation term for both paths is identical on the same target task, so the relative comparison reduces to the training term. To formalize this relationship and the systematic gains offered by indirect supervision, the next subsection presents a dominance criterion and its equivalent form, together with structural induction explanations.

\subsection{Dominance Criterion and Indirect Supervision Gains}
We compare the indirect and direct paths by model-structure induction and by transfer-risk scaling. The indirect path uses representations obtained from radar heart sounds via a heart-sound encoder and introduces an ECG decoder that mirrors the encoder for timestep-wise ECG reconstruction; perceptual consistency constraints may be added. The direct path relies on the heart-sound encoder and self-supervised projection heads to impose task signals without using ECG reconstruction supervision. This difference brings two main benefits: ECG decoding provides dense, low-noise, strongly inductive sequence-level supervision so each sequence of length $L$ yields $O(L)$ effective signals and the per-task sample size $n$ is amplified; ECG as a direct physiological representation broadens the coverage of task clusters over the feature space of the optimal shared representation $h^*$ and increases task diversity $\nu$. The indirect path thus constructs richer task clusters under diverse cardiac rhythms and morphologies, often making $t$ larger than in the direct path. Although the ECG decoder increases the task-specific complexity $C(\mathcal{F})$ during pretraining (affecting the training term), the systematic gains in $n$, $t$, and $\nu$ offset and surpass its impact on the risk upper bound.

\begin{definition}
To make the comparison precise, we isolate the contribution of multi-task training. Given a path characterized by $(n,t,\nu, C(\mathcal{F}))$, we define its training term as

\begin{equation}
\mathrm{TrainTerm}:=\frac{1}{\nu}\sqrt{\frac{C(\mathcal{H})+t\,C(\mathcal{F})}{n\,t}}.
\label{eq:TrainTermDef}
\end{equation}
\end{definition}

Under the scaling law in \eqref{eq:TLRiskScale}, two paths evaluated on the same target share the same adaptation term; consequently, path comparison reduces to the training term above. We parameterize the direct path as $(n_{\mathrm{dir}},t_{\mathrm{dir}},\nu_{\mathrm{dir}},C(\mathcal{F}_{\mathrm{dir}}))$ and the indirect path as $(n_{\mathrm{ind}},t_{\mathrm{ind}},\nu_{\mathrm{ind}},C(\mathcal{F}_{\mathrm{ind}}))$. With this notation, we say the indirect path dominates if $\mathrm{TrainTerm}_{\mathrm{ind}}<\mathrm{TrainTerm}_{\mathrm{dir}}$.

\begin{theorem}
A sufficient and explicit condition for the indirect path to dominate the direct path in terms of the training term is
\begin{equation}
\frac{1}{\nu_{\mathrm{ind}}}\sqrt{\frac{C(\mathcal{H})+t_{\mathrm{ind}}\,C(\mathcal{F}_{\mathrm{ind}})}{n_{\mathrm{ind}}\,t_{\mathrm{ind}}}}
<
\frac{1}{\nu_{\mathrm{dir}}}\sqrt{\frac{C(\mathcal{H})+t_{\mathrm{dir}}\,C(\mathcal{F}_{\mathrm{dir}})}{n_{\mathrm{dir}}\,t_{\mathrm{dir}}}}.
\label{eq:MainCriterion}
\end{equation}
\end{theorem}

By the definition of the training term, training-term dominance is exactly the comparison $\mathrm{TrainTerm}_{\mathrm{ind}}<\mathrm{TrainTerm}_{\mathrm{dir}}$, which expands to \eqref{eq:MainCriterion} after substituting the parameters of the two paths.

By \eqref{eq:MainCriterion}, in our task setting ECG reconstruction provides timestep-wise dense supervision, resulting in $n_{\mathrm{ind}}\!\gg\! n_{\mathrm{dir}}$; coverage across multiple rhythms, morphologies, and individuals yields $t_{\mathrm{ind}}\!\gtrsim\! t_{\mathrm{dir}}$ and $\nu_{\mathrm{ind}}\!\gg\!\nu_{\mathrm{dir}}$. Although introducing the ECG decoder may cause $C(\mathcal{F}_{\mathrm{ind}})$ to be slightly higher than $C(\mathcal{F}_{\mathrm{dir}})$, as long as the gains in $(n_{\mathrm{ind}},t_{\mathrm{ind}},\nu_{\mathrm{ind}})$ brought by dense supervision and task cluster expansion result in a smaller training term, the indirect path will achieve superiority in overall transfer learning risk. Intuitively, the indirect path trades ``strong supervision, strong diversity, strong coverage'' for ``moderate complexity growth'', and under the prior enhancement of ECG as a physiological dual representation, it can capture transferable features of heart sounds with lower variance and stronger shared structure, thus demonstrating systematic advantages.

\section{Experiments and Results}

\begin{table*}[!]
  \vspace{-3.0ex}
  \begin{center}
    \caption{Comparison of results between direct and indirect paths in five scenarios under different downstream tasks}
    \label{CP}

    {\scriptsize\bfseries Distortion-driven (IOU) $\uparrow$}\par\vspace{0.3ex}
    \setlength{\tabcolsep}{4pt}\renewcommand{\arraystretch}{1.05}
    \scriptsize
    \begin{tabularx}{\textwidth}{ccc|*{7}{>{\centering\arraybackslash}X}}
      \hline
      &  &  & \multicolumn{7}{c}{Distortion-driven(IOU) $\uparrow$} \\
      \cline{4-10}
      &  &  & \multicolumn{1}{c}{PR Interval IOU} & \multicolumn{1}{c}{QRS Interval IOU} & \multicolumn{1}{c}{ST Interval IOU} & \multicolumn{1}{c}{RR Interval IOU} & \multicolumn{1}{c}{PR Segment IOU} & \multicolumn{1}{c}{ST Segment IOU} & \multicolumn{1}{c}{Mean IOU} \\
      \hline

      \multirow{2}{*}{ID} & \multirow{2}{*}{Resting} & Direct Path   & 0.878 & 0.893 & 0.931 & 0.000280 & 0.775 & 0.897 & 0.729 \\
                           &                          & Indirect Path & \textbf{0.914} & \textbf{0.914} & \textbf{0.949} & \textbf{0.290}     & \textbf{0.814} & \textbf{0.916} & \textbf{0.800}   \\
      \hline

      \multirow{8}{*}{OOD} & \multirow{2}{*}{Valsalva} & Direct Path   & 0.777 & 0.791 & 0.845 & 0.000107 & 0.651 & 0.793 & 0.643 \\
                             &                           & Indirect Path & \textbf{0.802} & \textbf{0.807} & \textbf{0.858} & \textbf{0.217}    & \textbf{0.672} & \textbf{0.805} & \textbf{0.694} \\
      \cline{2-10}
                             & \multirow{2}{*}{Apnea}    & Direct Path   & 0.720 & 0.731 & 0.800 & 0.000161 & 0.544 & 0.724 & 0.586 \\
                             &                           & Indirect Path & \textbf{0.727} & \textbf{0.736} & \textbf{0.806} & \textbf{0.109}    & \textbf{0.555} & \textbf{0.732} & \textbf{0.611} \\
      \cline{2-10}
                             & \multirow{2}{*}{TiltUp}   & Direct Path   & \textbf{0.484} & \textbf{0.494} & \textbf{0.568} & 0.000    & \textbf{0.274} & \textbf{0.466} & \textbf{0.381} \\
                             &                           & Indirect Path & 0.430  & 0.441 & 0.516 & \textbf{0.0312}   & 0.249 & 0.425 & 0.349 \\
      \cline{2-10}
                             & \multirow{2}{*}{TiltDown} & Direct Path   & 0.755 & 0.766 & \textbf{0.827} & 6.78E-05 & 0.578 & \textbf{0.767} & 0.615 \\
                             &                           & Indirect Path & \textbf{0.767} & \textbf{0.768} & 0.825 & \textbf{0.114}    & \textbf{0.593} & 0.766 & \textbf{0.639} \\
      \hline
    \end{tabularx}

    \vspace{0.8ex}

    {\scriptsize\bfseries Perception-driven (ACC) $\uparrow$ and Transcendent (MAE) $\downarrow$}\par\vspace{0.3ex}
    \scriptsize
    \begin{tabularx}{\textwidth}{ccc|*{4}{>{\centering\arraybackslash}X}|>{\centering\arraybackslash}X}
      \hline
      &  &  & \multicolumn{4}{c}{Perception-driven(ACC) $\uparrow$} & \multicolumn{1}{c}{Transcendent (MAE) $\downarrow$} \\
      \cline{4-8}
      &  &  & \multicolumn{1}{c}{Subject Identification} & \multicolumn{1}{c}{BMI} & \multicolumn{1}{c}{Sex} & \multicolumn{1}{c}{Age} & \multicolumn{1}{c}{Blood Pressure} \\
      \hline

      \multirow{2}{*}{ID} & \multirow{2}{*}{Resting} & Direct Path   & 0.872  & 0.984 & \textbf{0.980} & 0.980 & 6.51 \\
                           &                          & Indirect Path & \textbf{0.989}  & \textbf{0.989} & 0.973 & \textbf{0.982} & \textbf{4.73} \\
      \hline

      \multirow{8}{*}{OOD} & \multirow{2}{*}{Valsalva} & Direct Path   & 0.331  & \textbf{0.735} & 0.642 & \textbf{0.551} & 15.2 \\
                             &                           & Indirect Path & \textbf{0.336}  & 0.644 & \textbf{0.659} & 0.538 & \textbf{12.1} \\
      \cline{2-8}
                             & \multirow{2}{*}{Apnea}    & Direct Path   & \textbf{0.0515} & 0.824 & \textbf{0.554} & \textbf{0.391} & 16.6 \\
                             &                           & Indirect Path & 0.0337 & \textbf{0.849} & 0.533 & 0.361 & \textbf{14.1} \\
      \cline{2-8}
                             & \multirow{2}{*}{TiltUp}   & Direct Path   & \textbf{0.0607} & 0.654 & 0.529 & \textbf{0.465} & 19.6 \\
                             &                           & Indirect Path & 0.046  & \textbf{0.894} & \textbf{0.571} & 0.326 & \textbf{14.8} \\
      \cline{2-8}
                             & \multirow{2}{*}{TiltDown} & Direct Path   & 0.250  & 0.774 & 0.597 & 0.563 & 14.9 \\
                             &                           & Indirect Path & \textbf{0.332}  & \textbf{0.845} & \textbf{0.629} & \textbf{0.594} & \textbf{13.4} \\
      \hline
    \end{tabularx}

  \end{center}
  \vspace{-0.0ex}
\end{table*}

\subsection{Datasets and Baseline}

\noindent \textbf{Datasets} The SciData Phase dataset \cite{schellenberger2020dataset} comprises clinically recorded radar vital sign data along with synchronized reference sensor signals, capturing a total of 86,459 seconds from thirty healthy individuals under five protocols: resting, Valsalva, apnea, tilt-up, and tilt-down. From this, we extract radar heart sound signals and ECGs as our experimental samples. The original radar heart sound and ECG signals are sampled at 2000 Hz; we downsample all signals to 250 Hz and apply filtering, denoising, and normalization. For each individual, we split the data into training and testing sets in an 8:2 ratio. In the training set, we employ a sliding window with a width of 2048 and a step size of 256 to segment continuous signal sequences for data augmentation, resulting in 29,298 training examples. In the testing set, continuous signal sequences are divided into non-overlapping segments of length 2048, resulting in 447 testing examples.

\noindent \textbf{Baseline} To provide a comprehensive comparison, we evaluate two canonical generative models, CVAE\cite{xia2023ecg} and CWGAN\cite{xia2023ecg}, alongside a distortion-driven masked generation model\cite{lee2024time} and a perception-driven contrastive learning model\cite{khosla2020supervised}.

\subsection{Distortion-driven downstream tasks}

We formulate a temporally position-sensitive ECG semantic segmentation task as a representative benchmark for distortion-dominated objectives, designed to probe the impact of reconstruction fidelity on boundary localization. Each ECG sequence is partitioned into four intervals and two segments, namely the PR interval, QRS duration, ST interval, and RR interval, together with the PR and ST segments. To accommodate the inherent physiological overlap between intervals and segments, we employ a six-class annotation scheme that permits label co-occurrence, enabling fine-grained characterization of waveform morphology and rhythmic boundaries. Using time-synchronized pairs of radar heart-sound and ECG recordings that are matched in length and sampling rate, we perform frame-wise six-class segmentation and compare the direct and indirect pathways under a unified evaluation protocol. Performance is assessed with the intersection-over-union (IoU) metric, and we report both per-class IoU and macro-averaged IoU to comprehensively quantify coverage and localization accuracy; results are summarized in Table \ref{CP}.

The indirect path consistently surpasses the direct path across all intervals and segments, with IoU gains of approximately 3–4 percentage points for PR/QRS/ST and a marked improvement for RR; the mean IoU increases from 0.729 to 0.8, indicating that dense supervision via ECG decoding enhances structural consistency and reduces segmentation error.

Compared with baselines, our method yields the best overall structural consistency, as summarized in Table \ref{CPOM}: mean IoU reaches 0.8, surpassing Supcom (0.770) and CVAE (0.677). PR/QRS/ST intervals and PR/ST segments improve by roughly 4--10 percentage points, while the RR interval improves from 0.193 to 0.290 relative to CVAE but remains below Supcom.

\begin{table*}[!]
  \vspace{-3.0ex}
  \begin{center}
    \caption{Comparison of results with other baseline methods}
    \label{CPOM}

    {\scriptsize\bfseries Distortion-driven (IOU) $\uparrow$}\par\vspace{0.3ex}
    \setlength{\tabcolsep}{4pt}\renewcommand{\arraystretch}{1.05}
    \scriptsize
    \begin{threeparttable}
    \begin{tabularx}{\textwidth}{c|*{7}{>{\centering\arraybackslash}X}}
      \hline
      & \multicolumn{7}{c}{Distortion-driven(IOU) $\uparrow$} \\
      \cline{2-8}
       & \multicolumn{1}{|c}{PR Interval IOU} & \multicolumn{1}{c}{QRS Interval IOU} & \multicolumn{1}{c}{ST Interval IOU} & \multicolumn{1}{c}{RR Interval IOU} & \multicolumn{1}{c}{PR Segment IOU} & \multicolumn{1}{c}{ST Segment IOU} & \multicolumn{1}{c}{Mean IOU} \\
      \hline

      CVAE   & 0.786 & 0.801 & 0.845 & 0.193 & 0.647 & 0.789 & 0.677 \\
      CWGAN  & --- & --- & ---&  --- &--- &--- &--- \\
      Masked & --- & --- & ---&  --- &--- &--- &---  \\
      Supcom & 0.809 & 0.874 & 0.898 & \textbf{0.445} & 0.714 & 0.877 & 0.770 \\
      Ours   & \textbf{0.914} & \textbf{0.914} & \textbf{0.949} & 0.290  & \textbf{0.814} & \textbf{0.916} & \textbf{0.800} \\
      \hline
    \end{tabularx}

    \vspace{0.8ex}

    {\centering
\scriptsize\bfseries Perception-driven (ACC) $\uparrow$ and Transcendent (MAE) $\downarrow$\par
}\vspace{0.3ex}
    
    \scriptsize
    \begin{tabularx}{\textwidth}{c|*{4}{>{\centering\arraybackslash}X}|>{\centering\arraybackslash}X}
      \hline
      & \multicolumn{4}{c}{Perception-driven(ACC) $\uparrow$} & \multicolumn{1}{c}{Transcendent (MAE) $\downarrow$} \\
      \cline{2-6}
       & \multicolumn{1}{c}{Subject Identification} & \multicolumn{1}{c}{BMI} & \multicolumn{1}{c}{Sex} & \multicolumn{1}{c}{Age} & \multicolumn{1}{c}{Blood Pressure} \\
      \hline

      CVAE   & 0.629  & 0.904 & 0.864 & 0.783 & 11.9 \\
      CWGAN  & 0.0579 & 0.897  & 0.581  & 0.411  & 14.3    \\
      Masked & 0.250  & 0.774  & 0.597  & 0.563  & 14.9    \\
      Supcom & \textbf{0.991}  & 0.987  & \textbf{0.973}  & 0.980   & 10.7    \\
      Ours   & 0.989  & \textbf{0.989}  & \textbf{0.973}  & \textbf{0.982}  & \textbf{4.73}    \\
      \hline
    \end{tabularx}
\begin{tablenotes}    
        \footnotesize              
        \item[*] “---” indicates that the model does not converge or generate reasonable results.
      \end{tablenotes}           
    \end{threeparttable}
  \end{center}
\end{table*}

\subsection{Perception-driven downstream tasks}

We establish four population-attribute–related classification tasks as perception-dominated downstream benchmarks, including subject identification, body mass index (BMI), sex, and age. These tasks are designed to assess the model’s capacity to preserve phenotype-discriminative representations, prioritizing distributional and semantic discernibility over pointwise amplitude fidelity. All labels are encoded in a one-hot format to align with a unified multi-class training and evaluation pipeline. Concretely, subject identification is formulated as a 30-way classification over individual participants; BMI is discretized into three classes using thresholds at 18.5 and 24.9; sex is treated as a binary classification; and age is partitioned into three classes with cutoffs at 24 and 30 years. Under the same training and validation protocol as above, we compare the direct and indirect pathways across these four tasks, and report results under a consistent evaluation protocol in Table \ref{CP}.

Overall, the indirect path achieves superior performance: subject identification improves from 0.872 to 0.989, BMI from 0.984 to 0.989, and age from 0.980 to 0.982, while sex remains comparable to the direct path. These findings suggest that ECG-aligned representations more effectively capture subject-specific variability while preserving discriminability for common demographic attributes.

Our method remains highly competitive against the strongest baseline (Table \ref{CPOM}): BMI and age are slightly higher than Supcom, sex is tied at 0.973, and subject identification is marginally lower (0.989 vs 0.991). Relative to CVAE/CWGAN/Masked, our method shows large gains (e.g., subject ID from 0.629/0.0579/0.250 to 0.989), indicating that ECG-aligned features better capture inter-subject differences while maintaining discrimination on population attributes.

\begin{figure*}[t]
\centering

\begin{subfigure}[t]{0.24\textwidth}
    \centering
    \begin{overpic}[width=\linewidth]{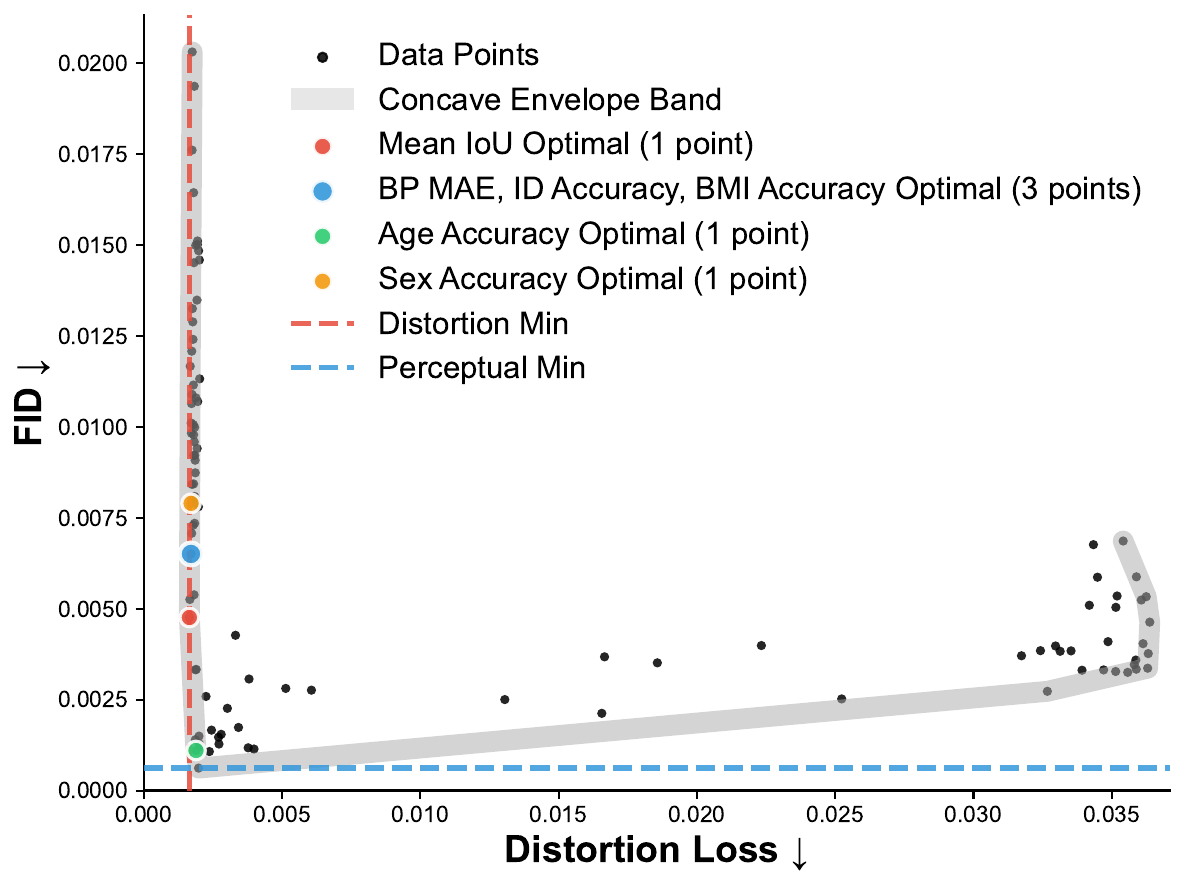}
    \end{overpic}
    \caption{} 
    \label{fig:three-rows:a}
\end{subfigure}\hfill
\begin{subfigure}[t]{0.24\textwidth}
    \centering
    \begin{overpic}[width=\linewidth]{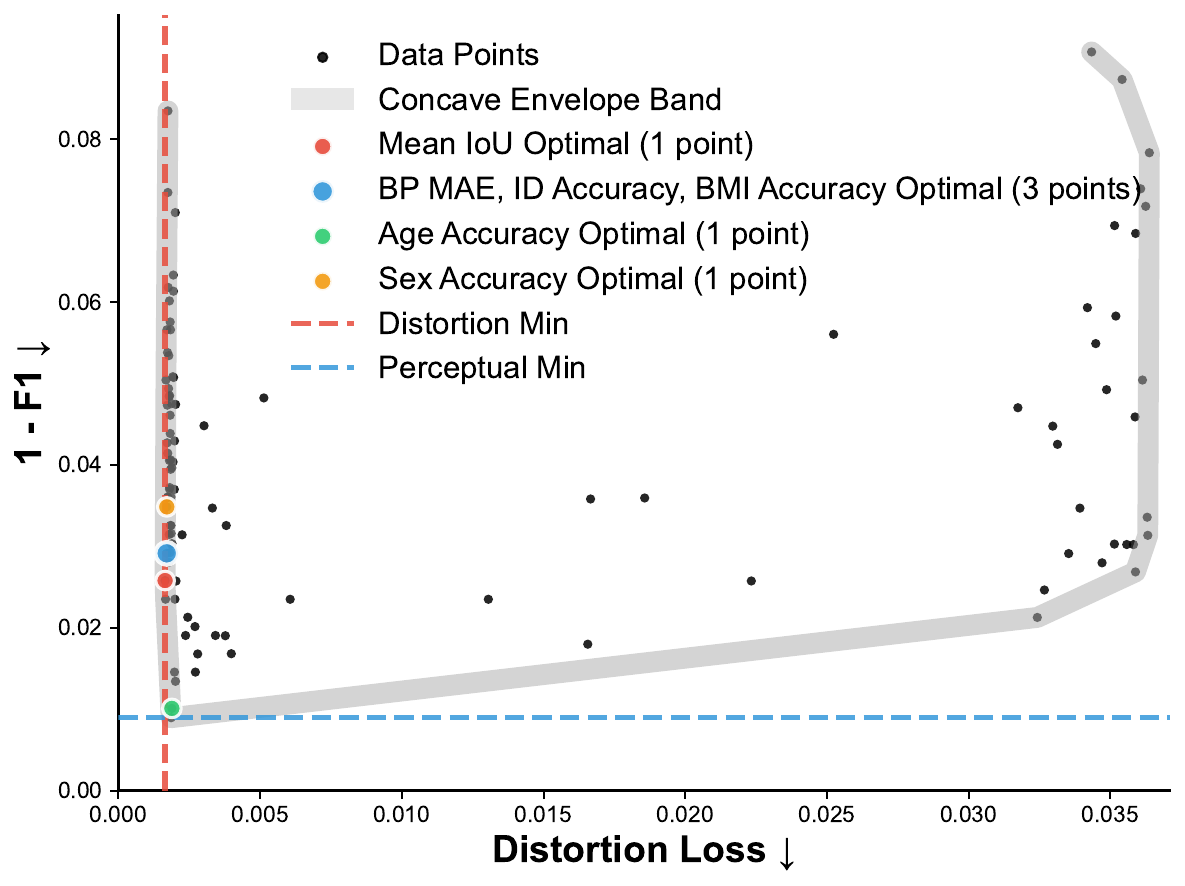}
    \end{overpic}
    \caption{}
    \label{fig:three-rows:b}
\end{subfigure}\hfill
\begin{subfigure}[t]{0.24\textwidth}
    \centering
    \begin{overpic}[width=\linewidth]{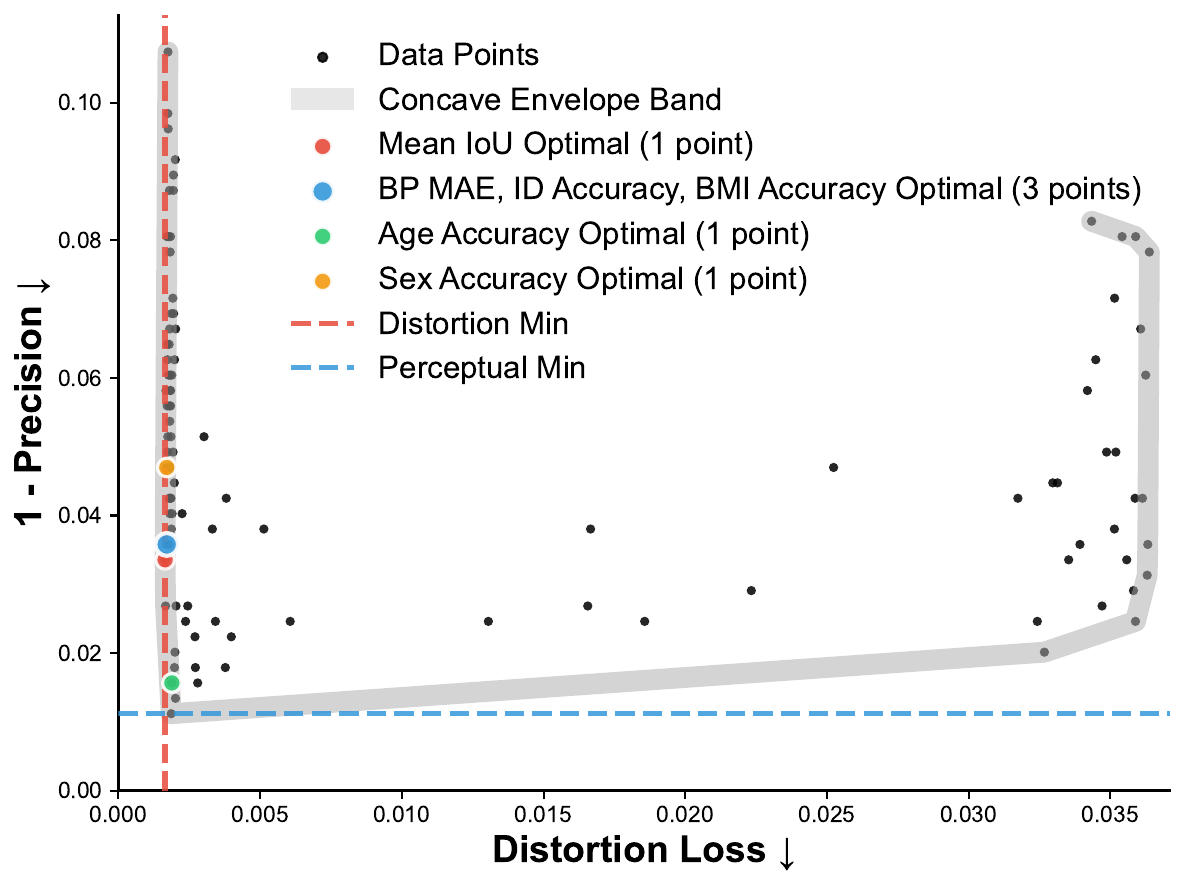}
    \end{overpic}
    \caption{}
    \label{fig:three-rows:c}
\end{subfigure}\hfill
\begin{subfigure}[t]{0.24\textwidth}
    \centering
    \begin{overpic}[width=\linewidth]{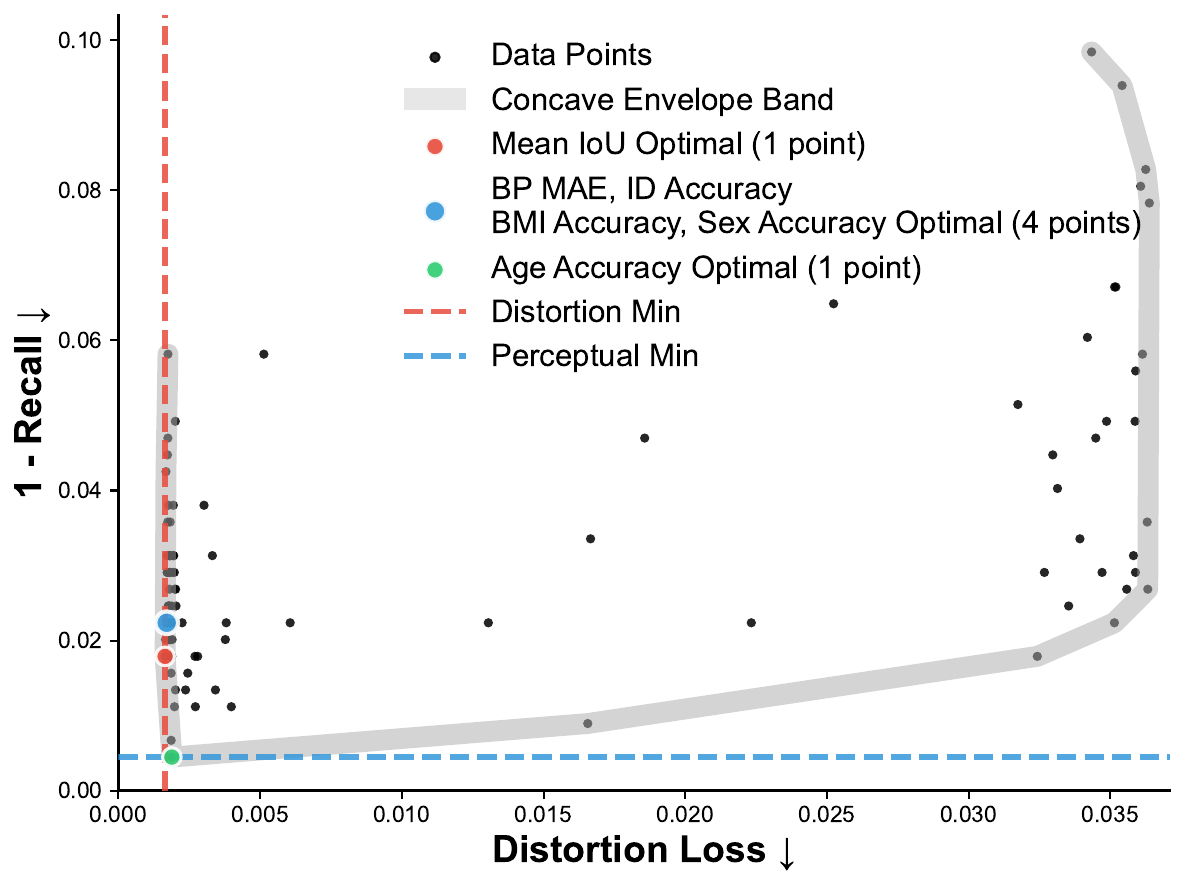}
    \end{overpic}
    \caption{}
    \label{fig:three-rows:d}
\end{subfigure}

\par\vspace{1.0ex} 

\begin{subfigure}[t]{0.24\textwidth}
    \centering
    \begin{overpic}[width=\linewidth]{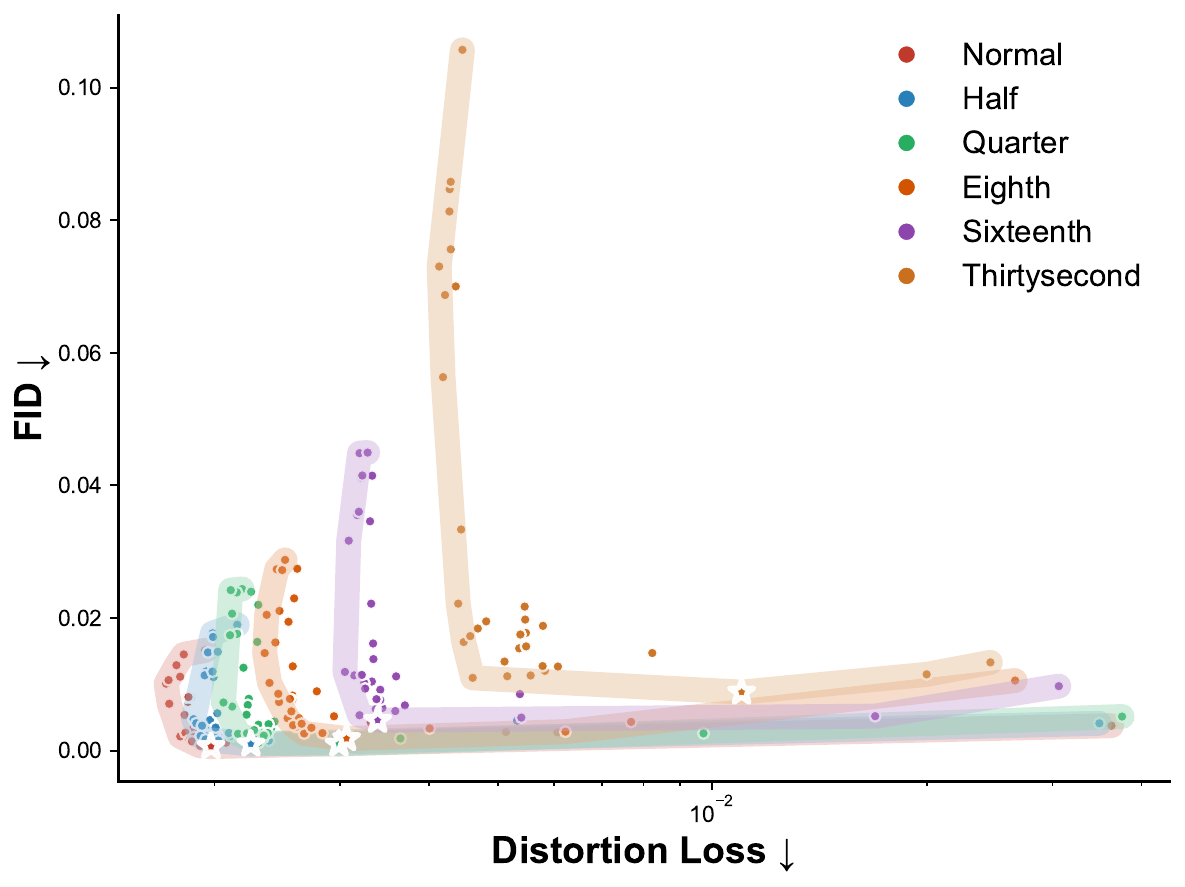}
    \end{overpic}
    \caption{}
    \label{fig:three-rows:e}
\end{subfigure}\hfill
\begin{subfigure}[t]{0.24\textwidth}
    \centering
    \begin{overpic}[width=\linewidth]{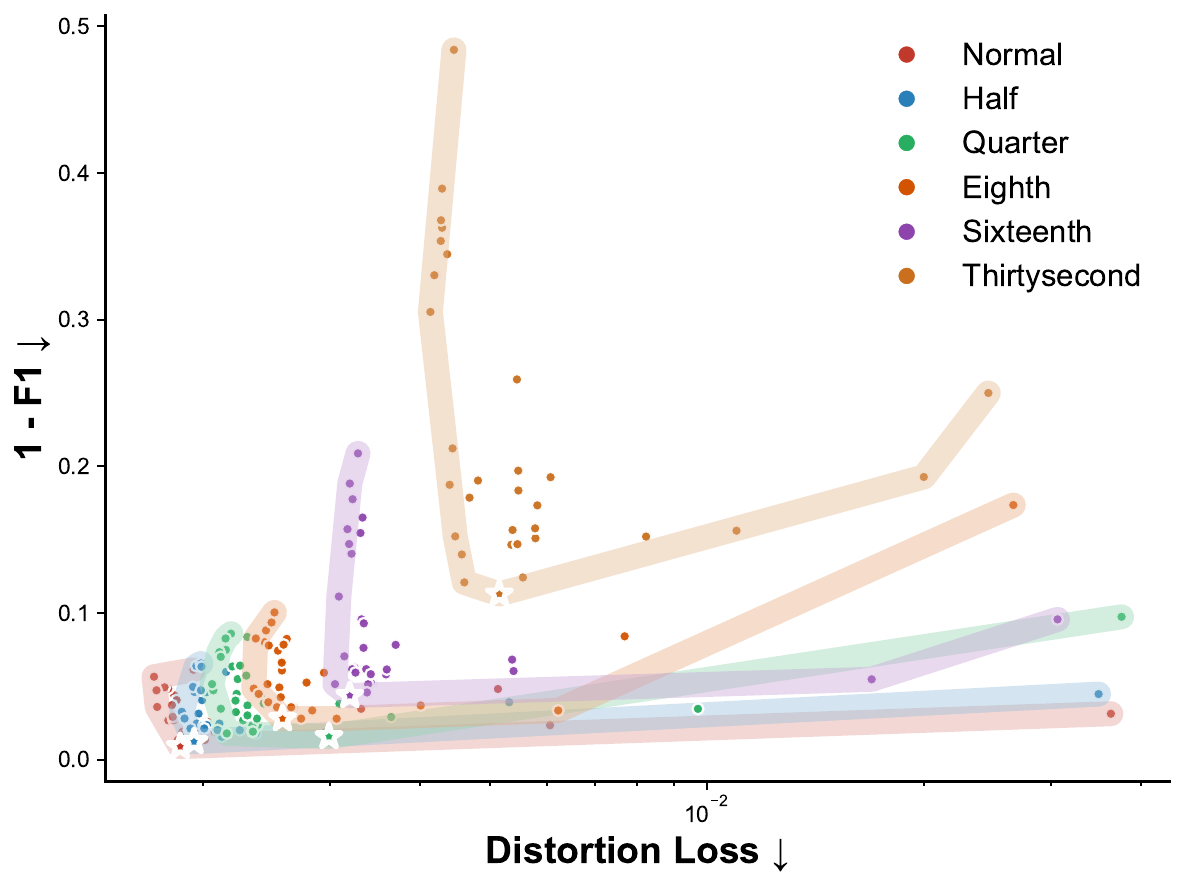}
    \end{overpic}
    \caption{}
    \label{fig:three-rows:f}
\end{subfigure}\hfill
\begin{subfigure}[t]{0.24\textwidth}
    \centering
    \begin{overpic}[width=\linewidth]{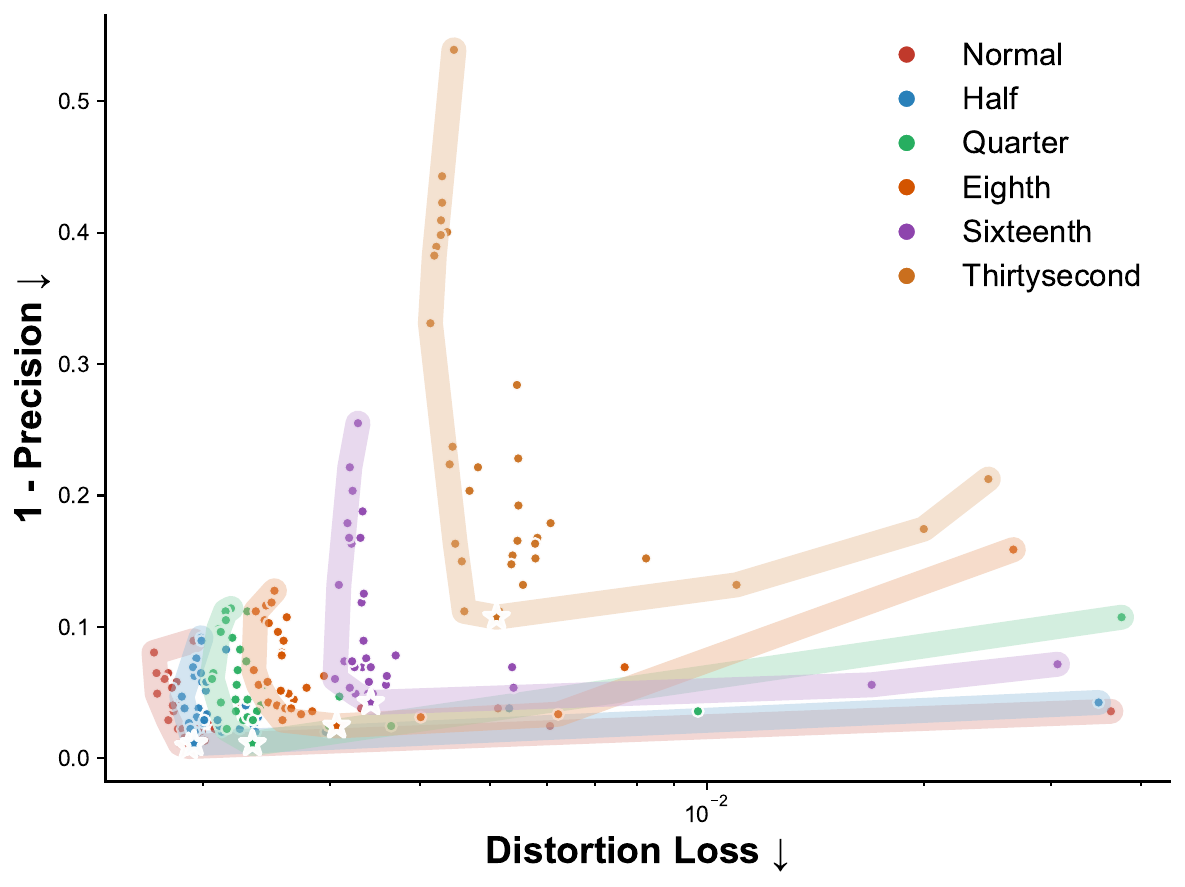}
    \end{overpic}
    \caption{}
    \label{fig:three-rows:g}
\end{subfigure}\hfill
\begin{subfigure}[t]{0.24\textwidth}
    \centering
    \begin{overpic}[width=\linewidth]{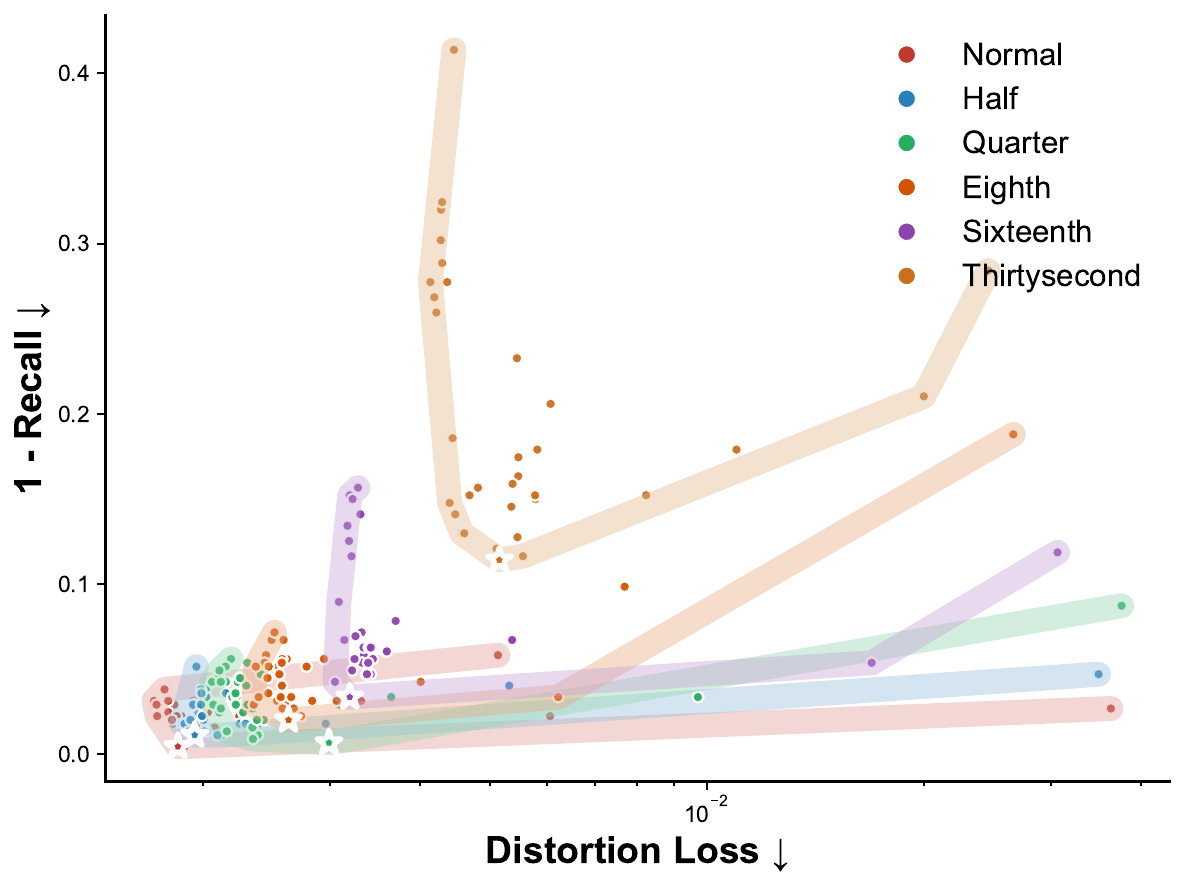}
    \end{overpic}
    \caption{}
    \label{fig:three-rows:h}
\end{subfigure}

\caption{(a–d) The relationships between the distortion metric and multiple perceptual metrics under varying combinations of $\lambda_d$ and $\lambda_p$, with the operating point corresponding to downstream optimal performance explicitly indicated. (e–h) Depict the relative positions and shapes of the scatter distributions between the distortion metric and each perceptual metric across different channel counts, that is, resolutions.}
\label{fig:fig4}

\end{figure*}

\subsection{Transcendent downstream tasks}

To transcend the limitations of distortion- and perception-dominated evaluations, we define a “transcendent” downstream task that aims to generate continuous arterial blood pressure (BP) waveforms from time-synchronized radar heart-sound and ECG signals matched in sequence length and sampling rate. This task foregrounds cross-modal physiological mapping and electromechanical coupling, rather than relying on numerical fidelity between synthetic and ground-truth ECG (the distortion axis) or on feature-level perceptual similarity (the perception axis); accordingly, performance on this task exhibits no monotonic or predictable relationship with either axis. During training and evaluation, BP is produced via frame-wise sequence-to-sequence regression, and performance is quantified using mean absolute error (MAE) as the primary metric to measure pointwise amplitude deviations. Within the same unified protocol as above, we can examine both the direct pathway—end-to-end BP generation from radar heart-sound—and the indirect pathway, which first synthesizes ECG from radar heart-sound and subsequently regresses BP from ECG. The results are reported in Table \ref{CP}.

The indirect path reduces MAE from 6.51 to 4.73, a relative reduction of approximately 27\%, indicating that ECG synthesis prior to regression better preserves BP-related dynamics and improves the accuracy and stability of cross-modal sequence modeling.

Our method achieves the lowest error (Table \ref{CPOM}): MAE 4.73 vs 10.7 (Supcom), 11.9 (CVAE), 14.3 (CWGAN), and 14.9 (Masked), a ~56\% reduction against the strongest baseline. This suggests that synthesizing ECG first and then regressing BP preserves dynamics relevant to arterial pressure, improving cross-modal sequence modeling.

\subsection{Scenario-driven out-of-distribution task evaluation}
To assess scenario-driven out-of-distribution generalization, Resting serves as the in-distribution baseline, while Valsalva, Apnea, Tilt up, and Tilt down are designated as the OOD test scenarios. Without altering preprocessing, windowing, or model capacity, we replicate the previously defined downstream task families on the OOD datasets and retain the same training and evaluation protocol and metrics as in the ID setting. We compare the direct and indirect pathways across the four scenarios and compute performance degradation normalized to the ID baseline for each task to quantify robustness under distribution shift. No fine-tuning is performed on OOD data to avoid information leakage. The results are shown in Table \ref{CP}.

Across the four scenarios, the indirect path exhibits stronger overall out-of-distribution generalization than the direct path. For distortion-driven tasks, it achieves higher mean IoU in three of the four scenarios and maintains a clear advantage on RR-interval IoU, although the direct path performs better on several metrics in TiltUp. For perception-driven tasks, the indirect path improves BMI in three scenarios, sex in three scenarios, and age in one scenario, while subject identification is mixed across conditions. For the transcendent task, the indirect path delivers markedly lower MAE in blood pressure estimation across all scenarios. Taken together, the indirect path shows smaller degradation on the overall task suite and better cross-task robustness under scenario-driven distribution shifts.

\section{Analysis and Discussions}

\begin{figure*}[h!]
\centering
  \includegraphics[width=1\textwidth]{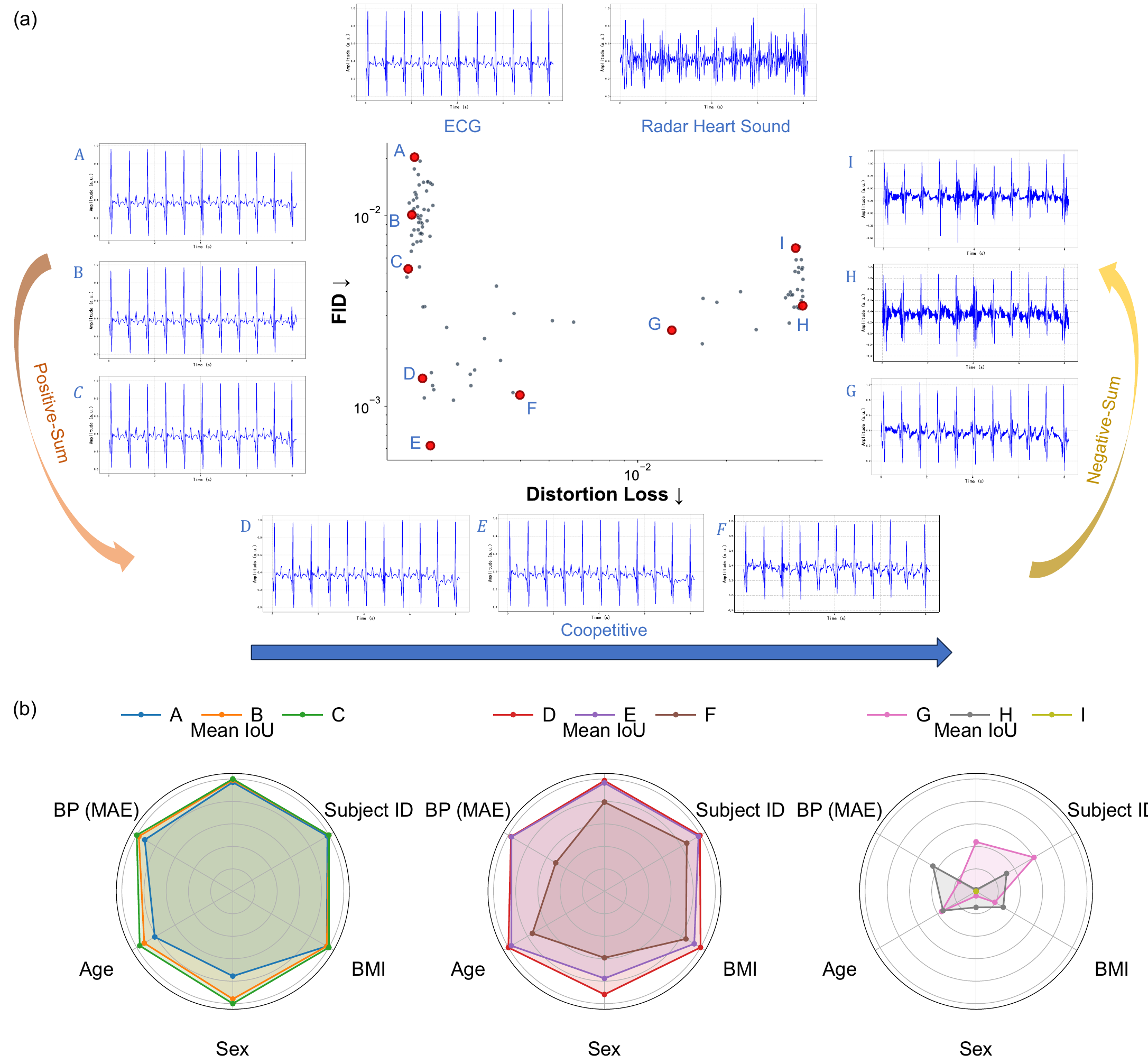}
  \vspace{-3mm} 
  \caption{(a) Visual comparison of representative points across the three stages; the loss weight configurations ($\lambda_d, \lambda_p$) for each point are A(700,1), B(5,1), C(1,9), D(1,60), E(1,80), F(1,2000), G(1,6000), H(1,800000), and I(1,170000). Note that all axes in the scatter plot are logarithmically scaled. (b) Radar plot of normalized evaluation metrics for these points, showing that downstream performance initially improves and then declines as the weights increase from A to I.}
  \vspace{-3mm}
  \label{Analysis}
\end{figure*}

\subsection{Mapping the Three-Phase Distortion–Perception Trade-off to the Downstream Optimum}

We investigate the relationship between the distortion-perception tradeoff and downstream optimality by conducting a controlled hyperparameter study of the indirect path. Specifically, we first fix $\lambda_p$ at 1 while varying $\lambda_d$ from 10,000 logarithmically to 1. Then, we fix $\lambda_d$ at 1 while varying $\lambda_p$ from 1 logarithmically to 10,000,000. This design ensures that the relative contribution of the perceptual loss to the overall objective function is monotonically increasing across both scans. Our analysis focuses on ECG generation to isolate the distortion-perception tradeoff. We use mean squared error (MSE) as the distortion metric and FID, precision, F1, and recall as perceptual metrics. The results are summarized in Figure \ref{fig:fig4}(\subref{fig:three-rows:a}–\subref{fig:three-rows:d}). To unify the optimality in the scatterplots, we plot 1-FID, 1-precision, 1-F1, and 1-recall as the perceptual axis.

Theoretical and empirical evidence consistently attribute this tradeoff to the properties of the three stages. Starting from the low-distortion region on the left side of the scatter plot, the point cloud converges toward the lower left corner, indicating that distortion dominates perceptual optimization in this region. As the trajectory approaches the distortion baseline, it reaches the first turning point between the first and second stages. After this turning point, the point cloud tilts downward to the right, indicating that the perceptual term becomes the primary driver and begins to reduce distortion. As the trajectory approaches the perceptual baseline, the second turning point marks the transition from the second to the third stage. After this turning point, the point cloud rapidly moves toward the upper right. At this point, overemphasizing perception not only fails to effectively guide the perceptual loss, but also further exacerbates the distortion loss, ultimately leading to failure in the generation task. In summary, the experiment reveals clear positive-sum, coopetitive, and negative-sum game stages in the distortion-perception game, as qualitatively shown in Figure \ref{Analysis}a. This figure intuitively illustrates the evolution of ECG generation from near-perfect synthesis to ultimate failure.

From the perspective of the relationship between the trade-off and downstream optima, the performance optimum for downstream tasks tends to reside in the vicinity of the coopetitive phase, as shown in Figure \ref{Analysis}b. In this region, the distortion term remains dominant and ensures numerical fidelity, while the perceptual term begins to impose constraints aligned with natural statistics and morphological priors; the two objectives exhibit limited but constructive synergy, thereby improving structural consistency relevant to task semantics without materially sacrificing reconstruction accuracy. As the perceptual weight continues to increase, the system transitions into a negative-sum phase: the model sacrifices lower-order precision in pursuit of perceptual similarity, resulting in rapid degradation of interval-/segment-level metrics and cross-task consistency. Accordingly, the emergence of the optimum within the coopetitive phase is underpinned by a minimally coupled allocation of the effective capacity of shared representations between lower-order recovery and higher-level semantics—an allocation that suppresses overfitting and non-realistic structures while avoiding task-level drift due to strong prior imposition, thus achieving a more desirable balance between distortion and perception.

\subsection{Resolution Effects on the Trade-off Curve}

Within the indirect-path framework, resolution is governed by the model’s global channel count: a larger number of channels affords greater expressive bandwidth and finer-grained structural representation, yielding higher effective resolution. To systematically assess how resolution shapes the distortion–perception trade-off, we reduce the baseline model’s channels to one-half, one-quarter, one-eighth, one-sixteenth, and one-thirty-second of the original configuration while holding all other training settings, hyperparameters, and data flows fixed, thereby isolating the effect of resolution alone. As shown in Figure \ref{fig:fig4}(\subref{fig:three-rows:e}–\subref{fig:three-rows:h}), progressive channel reduction shifts the scatter cloud away from the origin, indicating a concurrent deterioration of both distortion and perception metrics. Concurrently, the positive-sum region in the low-distortion regime contracts, the turning point shifts toward the lower right, and the coopetitive phase exhibits a flatter slope. These trends imply that higher perceptual weights are required to compensate for detail loss induced by limited capacity, yet this compensation more readily erodes low-order fidelity.

Mechanistically, reducing the number of channels contracts the effective capacity of the shared representations, attenuating the model’s ability to capture high-frequency morphology and long-range dependencies. Consequently, the perceptual term—grounded in natural statistics and morphological priors—struggles to maintain a stable synergy with numerical accuracy. In the positive-sum phase, the model cannot sustain morphology–rhythm coherence without incurring an increase in mean squared error. Upon entering the coopetitive phase, the perceptual term can partially correct structural defects, but its marginal gains diminish rapidly and it triggers adverse effects on the distortion term earlier. Before the onset of the negative-sum phase, the coopetitive region widens markedly, leading to a decline in peak performance and increased output uncertainty. These shifts directly affect downstream tasks: on the one hand, overall performance degrades as segmentation, detection, and classification metrics deteriorate in concert; on the other hand, identifying the optimum becomes more difficult—both because the widened, flatter coopetitive landscape reduces the effective signal-to-noise ratio (SNR), and because weakened cross-task consistency increases sensitivity to trade-off weights, inducing task-dependent shifts in the optimum. In summary, lowering resolution not only systematically raises the joint cost of distortion and perception, but also amplifies the search cost and uncertainty of downstream optima by compressing the effective capacity and transferability of the shared representations.

\section{Conclusion}
We presented TriDP-PTM, a pre-training model for contactless radar-based cardiac monitoring built on a dual-path comparative framework. By unifying ECG synthesis and downstream analysis through a multi-scale encoder-bottleneck-decoder and a feature discriminator, the model operationalizes the distortion–perception tradeoff and identifies a principled operating point that improves both clinical interpretability and task performance. Future work could focus on formalizing the generative–discriminative interplay that gives rise to the DP-tradeoff, enabling a unified paradigm with principled weight scheduling, cross-task generalization, and demonstrable clinical interpretability.

\printcredits

\bibliographystyle{cas-model2-names}

\bibliography{refs}





\end{document}